\documentclass[conference]{IEEEtran}
\usepackage[,bookmarks=true,colorlinks=true]{hyperref}
\usepackage[]{graphicx}
\usepackage[]{xcolor}

\usepackage[none]{hyphenat}
\usepackage{algorithm}
\usepackage{algorithmic}
\usepackage{amsthm}
\usepackage{ifthen}
\usepackage{subfig}
\usepackage[export]{adjustbox}
\usepackage{here}
\usepackage{multirow}
\usepackage{multicol}
\usepackage{url}
\usepackage{dblfloatfix}

\begin{document}

\renewcommand{\figurename}{Fig.}
\renewcommand{\tablename}{Table}

\newcommand{\cudnn}{cuDNN}
\newcommand{\ucudnn}{$\mu$-cuDNN}
\newcommand{\wdiv}{WD}
\newcommand{\wreuse}{WR}

\newcommand{\kfc}{TSUBAME-KFC/DL}
\newcommand{\tthree}{TSUBAME 3}
\newcommand{\dgxone}{DGX-1}

\newcommand{\caffe}{Caffe}
\newcommand{\nvcaffe}{NVCaffe}
\newcommand{\caffetwo}{Caffe2}
\newcommand{\tensorflow}{TensorFlow}

\newcommand{\keighty}{K80}
\newcommand{\phundred}{P100-SXM2}
\newcommand{\vhundred}{V100-SXM2}

\newcommand{\handle}{\texttt{cudnnHandle\_t}}
\newcommand{\uhandle}{\texttt{UcudnnHandle\_t}}

\newcommand{\ttfwd}{\texttt{Forward}}
\newcommand{\ttbwddata}{\texttt{BackwardData}}
\newcommand{\ttbwdfilter}{\texttt{BackwardFilter}}

\newcommand{\ttall}{\texttt{all}}
\newcommand{\ttpoweroftwo}{\texttt{powerOfTwo}}
\newcommand{\ttundivided}{\texttt{undivided}}

\newcommand{\argmin}{\mathop{\rm argmin}\limits}
\newcommand{\kernelset}{\mathcal{K}}

\newcommand{\alexnet}{AlexNet}
\newcommand{\resnet}{ResNet}
\newcommand{\densenet}{DenseNet}

\newcommand{\todo}[1]{\textcolor{red}{TODO: {#1} }}
\newcommand{\bhline}{\noalign{\hrule height 1.25pt}}
\newcommand{\halffigwidth}{0.5\linewidth}
\newcommand{\figwidth}{\linewidth}
\newcommand{\smallfigwidth}{0.4\textwidth}

\newcommand{\figref}[1]{\figurename \ \ref{#1}}
\newcommand{\tabref}[1]{\tablename \ \ref{#1}}
\newcommand{\algoref}[1]{Algorithm \ref{#1}}

\input{performance.tex}

\IEEEoverridecommandlockouts

\title{\ucudnn: Accelerating Deep Learning Frameworks with Micro-Batching}
\author{
  \IEEEauthorblockN{
    Yosuke Oyama\IEEEauthorrefmark{1},
    Tal Ben-Nun\IEEEauthorrefmark{2},
    Torsten Hoefler\IEEEauthorrefmark{2},
    Satoshi Matsuoka\IEEEauthorrefmark{3}\,\IEEEauthorrefmark{1}
  }
  \IEEEauthorblockA{
    \IEEEauthorrefmark{1}Department of Mathematical and Computing Science,
    Tokyo Institute of Technology,
    Tokyo, Japan\\
    \IEEEauthorrefmark{2}Department of Computer Science,
    ETH Zurich,
    Zurich, Switzerland\\
    \IEEEauthorrefmark{3}RIKEN Center for Computational Science,
    Hyogo, Japan\\
    \texttt{oyama.y.aa@m.titech.ac.jp},
    \texttt{\{talbn,htor\}@inf.ethz.ch},
    \texttt{matsu@acm.org}
  }
}


\maketitle

\begin{abstract}
NVIDIA \cudnn \ is a low-level library that provides GPU kernels frequently used in deep learning.
Specifically, \cudnn \ implements several equivalent convolution algorithms, whose performance and memory footprint may vary considerably, depending on the layer dimensions.
When an algorithm is automatically selected by \cudnn, the decision is performed on a per-layer basis, and thus it often resorts to slower algorithms that fit the workspace size constraints.
We present \ucudnn, a transparent wrapper library for \cudnn, which divides layers' mini-batch computation into several micro-batches.
Based on Dynamic Programming and Integer Linear Programming, \ucudnn \ enables faster algorithms by decreasing the workspace requirements.
At the same time, \ucudnn \ keeps the computational semantics unchanged, so that it decouples statistical efficiency from the hardware efficiency safely.
We demonstrate the effectiveness of \ucudnn \ over two frameworks, \caffe \ and \tensorflow,
achieving speedups of \perf{caffe-time-alexnet-p100-64mb-speedup-conv}x for \alexnet \ and \perf{caffe-time-resnet18-p100-64mb-speedup-conv}x for \resnet-18 on \phundred \ GPU.
These results indicate that using micro-batches can seamlessly increase the performance of deep learning, while maintaining the same memory footprint.
\end{abstract}

\section{Introduction} \label{section:introduction}

Prevalent Deep Neural Networks (DNNs) are becoming increasingly deeper and are trained with large batch sizes.
Specifically, state-of-the-art DNNs contain hundreds of layers \cite{Krizhevsky2012,He2016},
and utilize batch sizes in the order of thousands \cite{Goyal2017,2017arXiv171104325A,2017arXiv171100489S}.

Large batches are also favored by distributed data-parallel deep learning frameworks,
because they improve utilization of accelerators, as well as hiding the communication of parameter gradients in the computation efficiently.
Consequently, the batch size per accelerator (e.g., GPU) should be large to achieve better scaling.
Since the memory usage of a DNN is nearly proportional to the layer size and the batch size,
the accelerator memory tends to be used at full capacity in most real-world cases.

\begin{figure*}[t]
  \centering
  \subfloat[Execution time of all layers.]{
    \includegraphics[width=0.5\linewidth,valign=t]{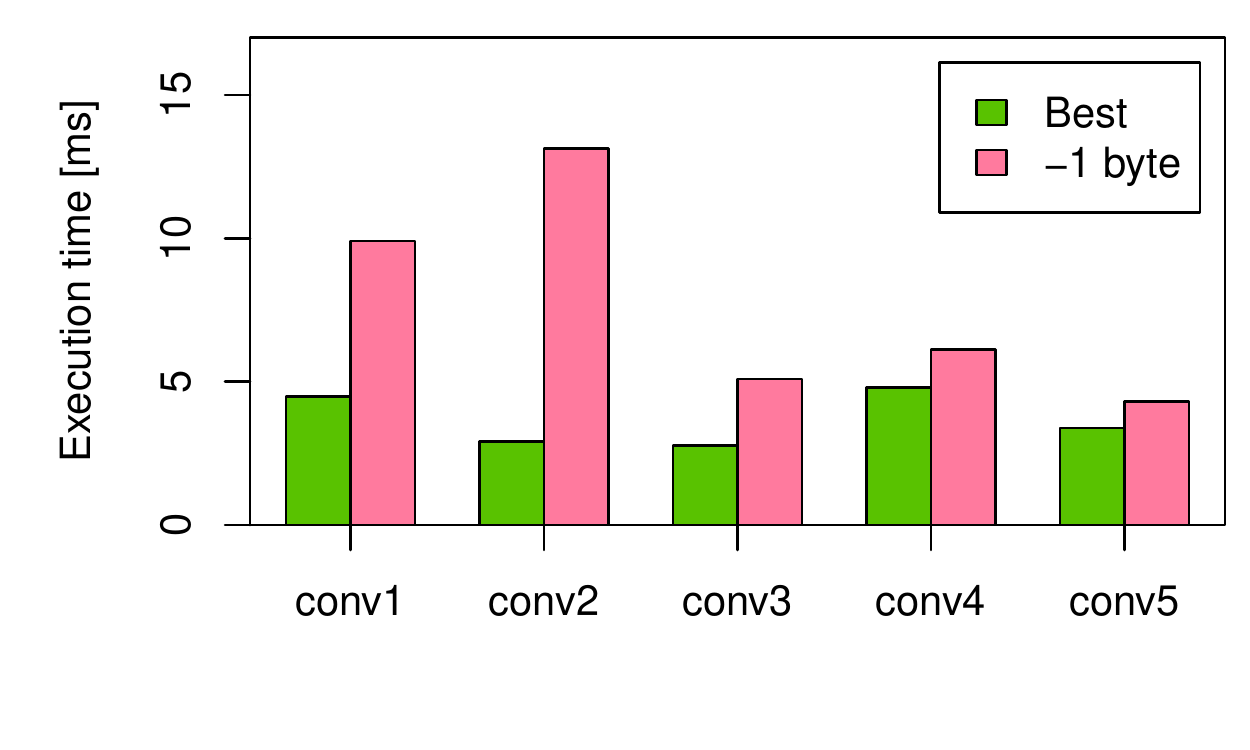}
  }
  \subfloat[Execution time vs. execution time of conv2. {\large $\circ$} and {\large $\diamond$} represent the ``Best'' and the ``-1 byte'' respectively.]{
    \includegraphics[width=0.5\linewidth,valign=t]{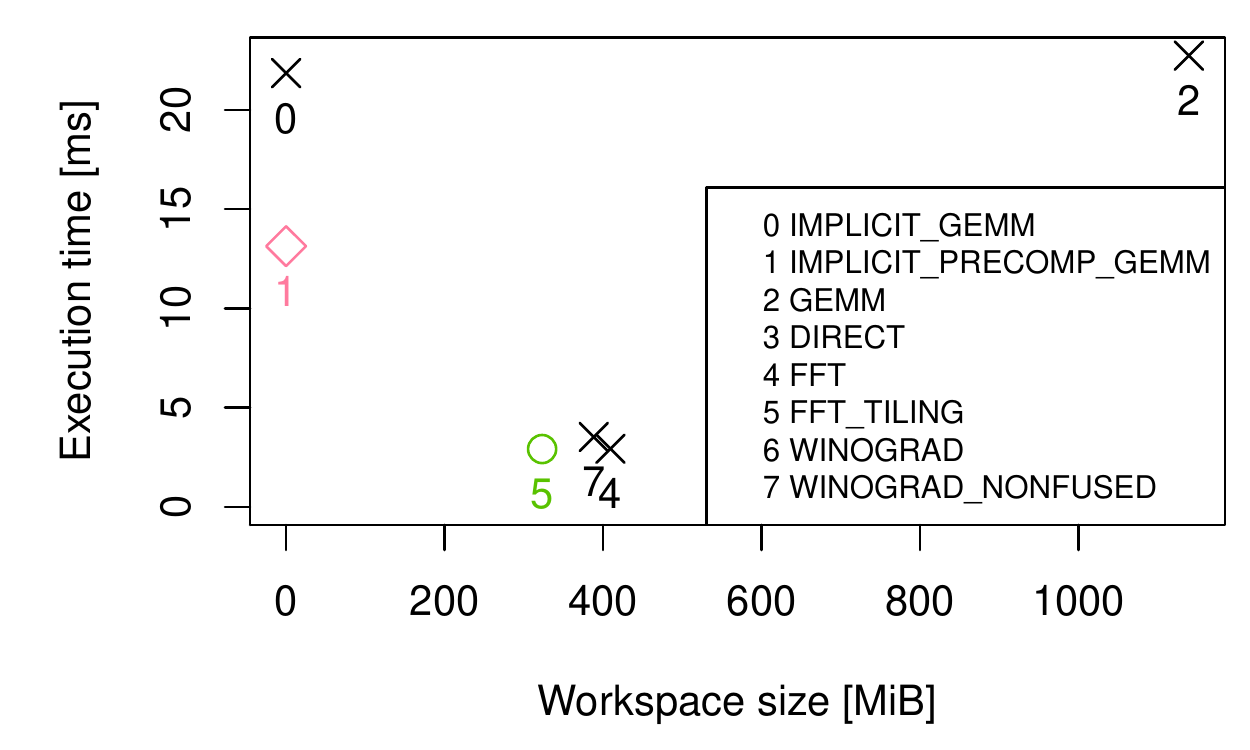}
  }
  \caption{Execution time of \cudnn \ 7.0.1 forward convolution of single-column \alexnet \ \cite{one_wired_trick} with different workspace sizes.
    The ``Best'' case always chooses the fastest algorithm regardless of workspace size, while in the ``-1 byte'' case the maximum workspace size is limited to 1 byte less than the best algorithm.}
  \label{figure:alexnet_1byte_less_ws}
\end{figure*}

This ``limited memory scenario'' is also exhibited in \cudnn \  \cite{cudnn}, a deep learning kernel library for NVIDIA GPUs.
\cudnn \ provides a variety of computational primitives for deep neural networks, and is widely used in deep learning frameworks,
such as Caffe \cite{jia2014caffe} and others \cite{tensorflow2015-whitepaper,2016arXiv160502688short,chainer_learningsys2015}.
\cudnn \ provides up to eight different algorithms to perform convolutions, each of which requires different temporary storage (workspace) schemes.
To guide users to determine the best algorithm for a given maximum workspace size,
\cudnn \ provides a function \texttt{cudnnGetConvolution*Algorithm} (\texttt{*} is one of convolution types, \texttt{Forward}, \texttt{BackwardData} and \texttt{BackwardFilter}),
that benchmarks all the algorithms and chooses the best algorithm, either with respect to computation time or memory usage.
However, if the workspace size requested by a fast algorithm is one byte larger than provided, \cudnn \ will resort to a slower algorithm that requires less workspace.
In fact, the performance impact can be 4.51x in the 2nd convolutional layer of \alexnet, as shown in \figref{figure:alexnet_1byte_less_ws}.


In this paper, we propose \ucudnn, a transparent wrapper for \cudnn \ that attempts to mitigate the aforementioned inefficiency.
In order to utilize fast convolution algorithms with limited size of workspace,
\ucudnn \ automatically divides layer mini-batch computation into several micro-batches and perform multiple convolutions sequentially.
\ucudnn \ decouples the statistical efficiency (speed of accuracy/loss improvement with fixed amount of parameter updates)
from the hardware efficiency (speed of computations with fixed amount of parameter updates), improving only the latter.
Using micro-batches, \ucudnn \ improves the utilization of the accelerators without incurring any reduction in training accuracy.

The contributions of this paper are as follows:
\begin{itemize}
\item We present a method to automatically divide mini-batch training into several ``micro-batches'', so that faster algorithms are utilized with tight workspace constraints.
\item We propose two different workspace allocation policies, which enable optimization of multiple convolutional layers with inter-dependencies.
\item We evaluate \ucudnn \ over two different deep learning frameworks, \caffe \ and \tensorflow, showing that it can mitigate the inefficiency of \cudnn \, even with state-of-the-art Convolutional Neural Networks (CNNs), such as \alexnet \ and \resnet.
\end{itemize}

\section{The Anatomy of Convolutional Neural Networks}
\begin{figure}[t]
  \centering
  \includegraphics[width=\smallfigwidth]{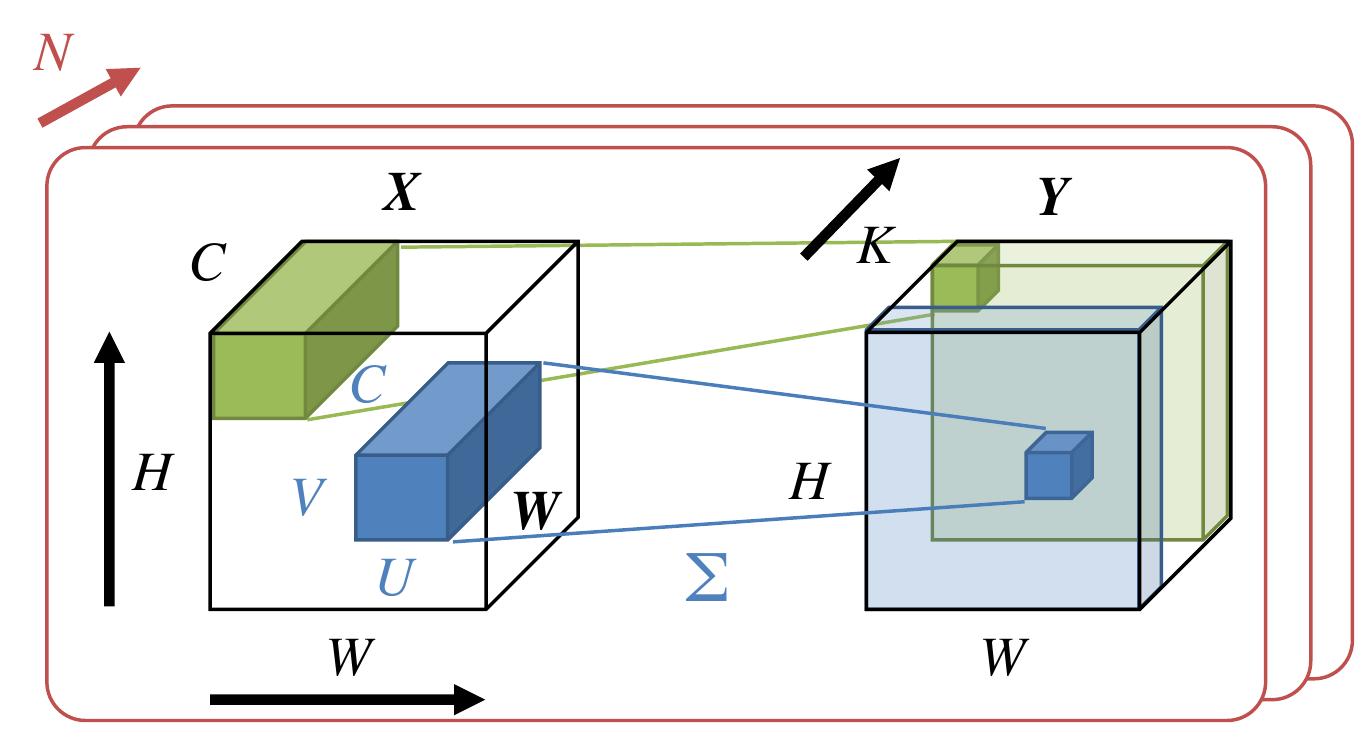}
  \caption{Two-dimensional convolution. Each element of $\mathbf{Y}$ is set to be a sum of element-wise products between partial $C \times V \times U$ area of $\mathbf{X}$ and one filter from $\mathbf{W}$.}
  \label{fig:conv}
\end{figure}

Convolution operations in Convolutional Neural Networks (CNNs) apply multiple filters to a batch of channels of two-dimensional data (Algorithm \ref{algo:conv}, \figref{fig:conv}).
In particular, input and output tensors are represented as four-dimensional tensors with dimensions ($N,C,H,W$), where $N$ is the mini-batch size, $C$ is the number of channels, and $H$ and $W$ represent image height and width, respectively.
Similarly, the filter tensor is represented as four-dimensional ($K,C,V,U$) tensor, where $K$ is the number of output channels and $V,U$ represent kernel height and width.

\begin{algorithm}[t]
  \caption{Pseudo-code of two-dimensional convolution.}
  \label{algo:conv}
  \begin{algorithmic}[1]
    \STATE \makebox[4.5cm][l]{for($n = 0$; $n < N$; $n$++)} // Mini-batch loop
    \STATE \makebox[4.5cm][l]{\ for($k = 0$; $k < K$; $k$++)} // Output channel loop
    \STATE \makebox[4.5cm][l]{\ \ for($h = 0$; $h < H$; $h$++)} // Height loop
    \STATE \makebox[4.5cm][l]{\ \ \ for($w = 0$; $w < W$; $w$++)} // Width loop
    \STATE \makebox[4.5cm][l]{\ \ \ \ for($c = 0$; $c < C$; $c$++)} // Input channel loop
    \STATE \makebox[4.5cm][l]{\ \ \ \ \ for($v = 0$; $v < V$; $v$++)} // Kernel width loop
    \STATE \makebox[4.5cm][l]{\ \ \ \ \ \ for($u = 0$; $u < U$; $u$++)} // Kernel height loop
    \STATE \ \ \ \ \ \ \ {\small $\mathbf{Y}[n,k,h,w]$ += $\mathbf{W}[k,c,v,u] \times \mathbf{X}[n,c,h+v,w+u]$;}
  \end{algorithmic}
\end{algorithm}

The two-dimensional convolution is composed of seven-nested loops (\algoref{algo:conv}).
The innermost three loops compute the actual convolution, where one element of the input tensor $\mathbf{X}$ is multiplied and accumulated to one element of the output tensor $\mathbf{Y}$.
The remaining loops iterate over all elements of $\mathbf{Y}$.
The key observation is that in order to solve the problem described in {Section \ref{section:introduction}}, there is no dependency inside the mini-batch loop between different iterations.
This is intuitive because in training or inference we compute parameter gradients or outputs with respect to different data samples, so this is equivalent to computing $N$ different CNNs concurrently.
This observation motivates us to apply loop tiling to the mini-batch loop, so that we can reduce the resident workspace size.

The only exception to the inter-sample independency is the computation of parameter gradients;
\begin{eqnarray}
\frac{\partial L}{\partial \mathbf{W}} = \frac{1}{N} \sum_{n=1}^{N} \frac{\partial L_n}{\partial \mathbf{Y}_n} * \mathbf{X}_n \nonumber ,
\end{eqnarray}
where $L$ and $L_n$ is the loss function with respect to a mini-batch and a sample $n$ respectively, and $*$ is the convolution operation \cite{b70224ae04784e15b91d2056c46924a6}.
The semantics of this computation is, however, not violated by the loop splitting, only if each of the iterations is performed sequentially.

In \cudnn, there are three operations related to the two-dimensional convolution;
\ttfwd \ for forward computation (\figref{fig:conv}), \ttbwddata \ for computing neuron errors in back-propagation, \ttbwdfilter \ for computing parameter gradients in back-propagation.

Although \ttfwd \ and \ttbwddata \ can directly be divided into several micro-batches, \ttbwdfilter \ cannot, since there are output dependencies on the accumulated parameter gradients tensor $\mathbf{dW}$.
However, we can still divide the loops by running \ttbwdfilter \ multiple times while accumulating the results, i.e., output scale $=1$ in \cudnn.
Therefore, loop splitting can be achieved by repeating \cudnn \ kernels one or more times for any convolution-related operation, regardless of the underlying method.

\section{\ucudnn}
\begin{figure}[t]
  \centering
  \includegraphics[width=\figwidth]{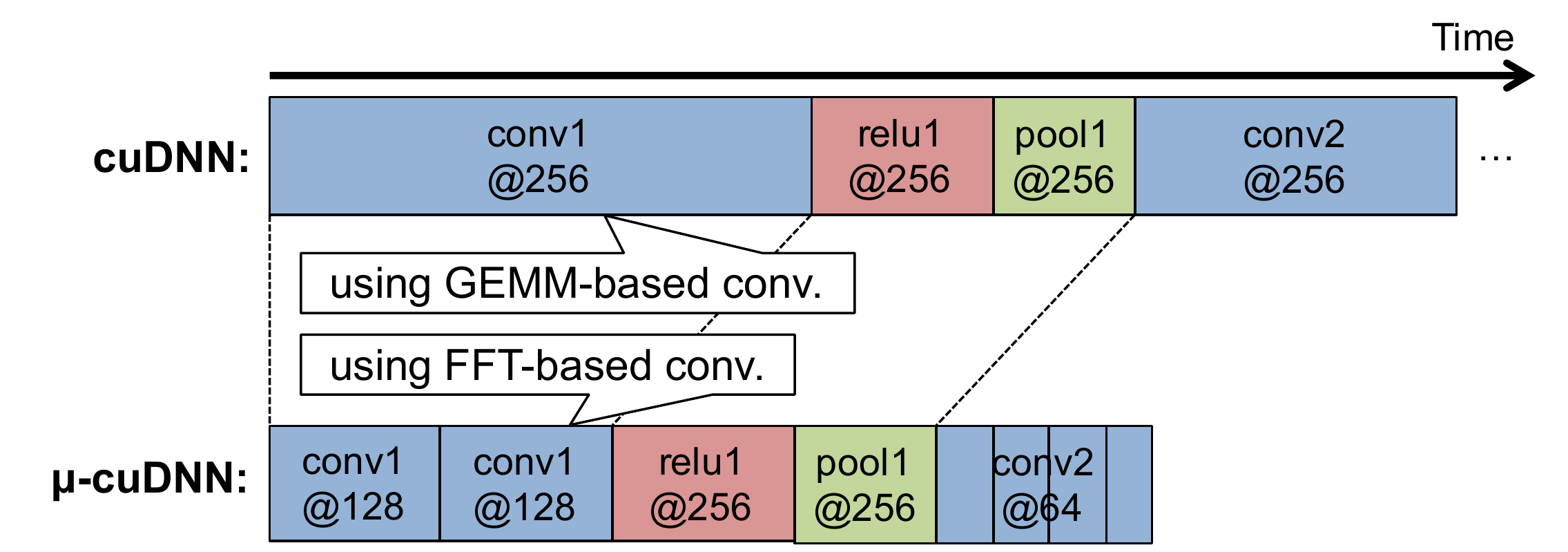}
  \caption{The conceptual timeline of \ucudnn.
    ``@256'' means that each computation is executed with batch-size of 256.
    \ucudnn \ splits one convolution operation into one or more disjoint subsets of the mini-batch.
  }
  \label{figure:ucudnn}
\end{figure}

\ucudnn \ is a transparent C++ wrapper library for \cudnn, which can easily be integrated into most deep learning frameworks \cite{jia2014caffe,caffe2,tensorflow2015-whitepaper,chainer_learningsys2015}.
The key concept of \ucudnn \ is that it automatically divides a mini-batch to several batches (referred to as ``micro-batches'' in this paper) and optimizes their sizes, to utilize faster convolution algorithms (\figref{figure:ucudnn}).

\subsection{\ucudnn \ Methodology} \label{subsection:ucudnn_methodology}
\begin{figure}[t]
  \centering
  \includegraphics[width=\figwidth]{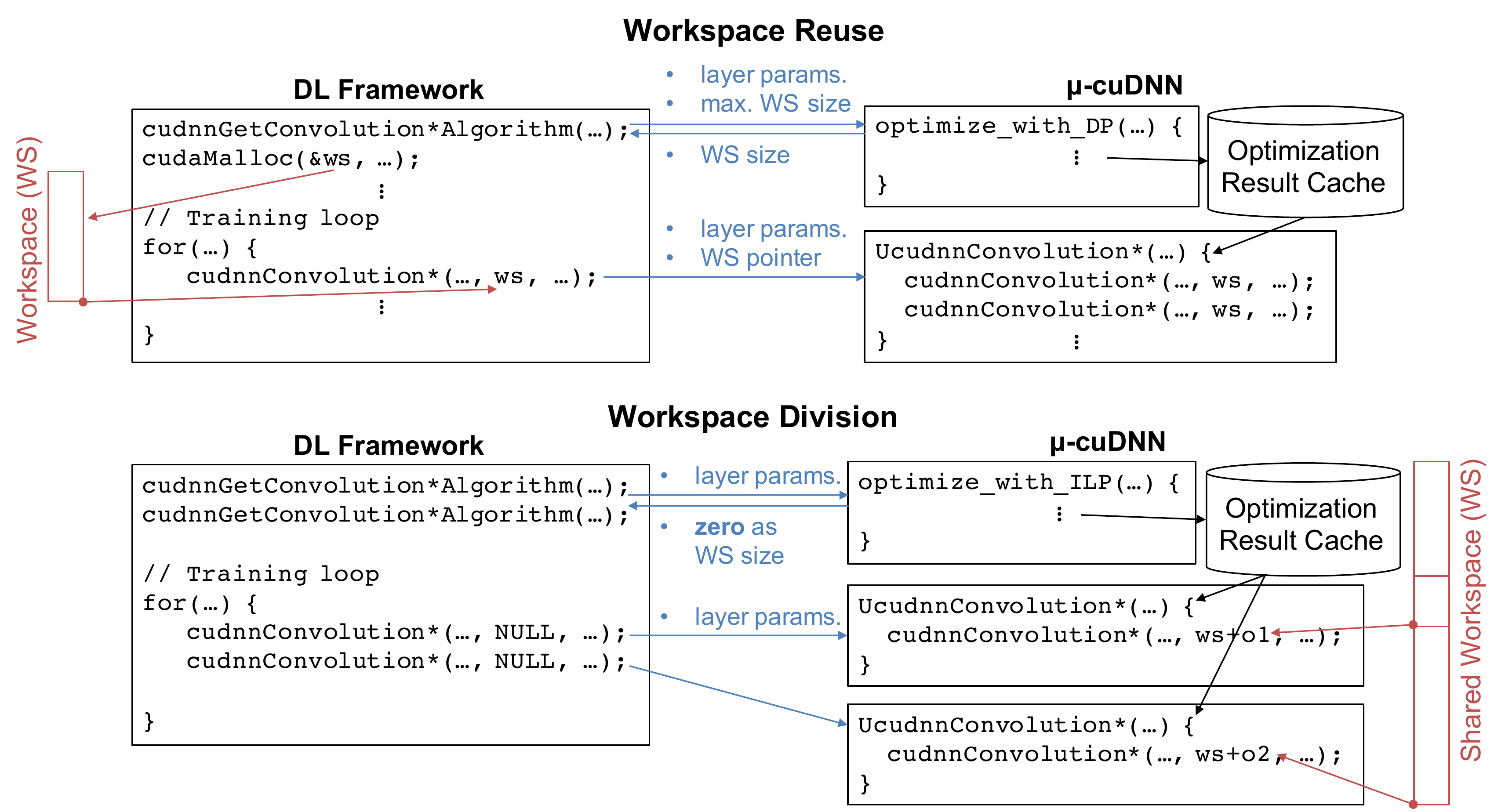}
  \caption{Overview of \wreuse \ and \wdiv. \ucudnn \ optimizes micro-batch sizes and internally calls \cudnn \ functions, via the \cudnn \ interfaces.}
  \label{figure:wrwd}
\end{figure}

\ucudnn \ library employs one of two workspace utilization policies to optimize micro-batches for convolution kernels (\figref{figure:wrwd}):

\begin{itemize}
\item {\bf Workspace Reuse (\wreuse)}: \wreuse \ allocates one workspace per layer, sharing the space between the internal micro-batches.
  In this scheme, each layer is assumed to use the workspace exclusively, hence the total size of the workspaces is in proportion to the number of convolutional layers.
\item {\bf Workspace Division (\wdiv)}: \wdiv \ allocates one workspace per network, and assigns different segments to each convolutional layer.
  \wdiv \ enables small groups of convolution operations, as in the Inception module \cite{going_deeper_with}, to run concurrently with larger workspaces.
  In \wdiv, the actual workspace is managed by \ucudnn \ rather than the deep learning framework.
  This is because conventional frameworks allocate each workspace separately, lacking a global view of the entire network's workspace requirements.
\end{itemize}

\wreuse \ and \wdiv \ both rely on the parameters of one or more convolution kernel(s), the mini-batch size, and the maximum workspace size.
The output of \ucudnn \ is a division of the mini-batch, and ``micro-configurations''; a pair of a convolution algorithm and micro-batch size for each convolution micro-batch.
In this paper, we define ``configuration'' of a segmented convolution kernel as ``a list of micro-configurations''.
For example, if a kernel with a mini-batch size of 256 is equally divided into four micro-batches and each of them uses algorithm $X$, the configuration is represented as $\{(X, 64), (X, 64), (X, 64), (X, 64)\}$.
Also we define concatenation of two lists as $+$, such as $\{a,b\} + \{c,d\} = \{a,b,c,d\}$ and $\{a\} + \emptyset = \{a\}$.

\subsection{\wreuse \ Algorithm} \label{subsection:wr_algorithm}
\begin{figure}[t]
  \centering
  \includegraphics[width=\figwidth]{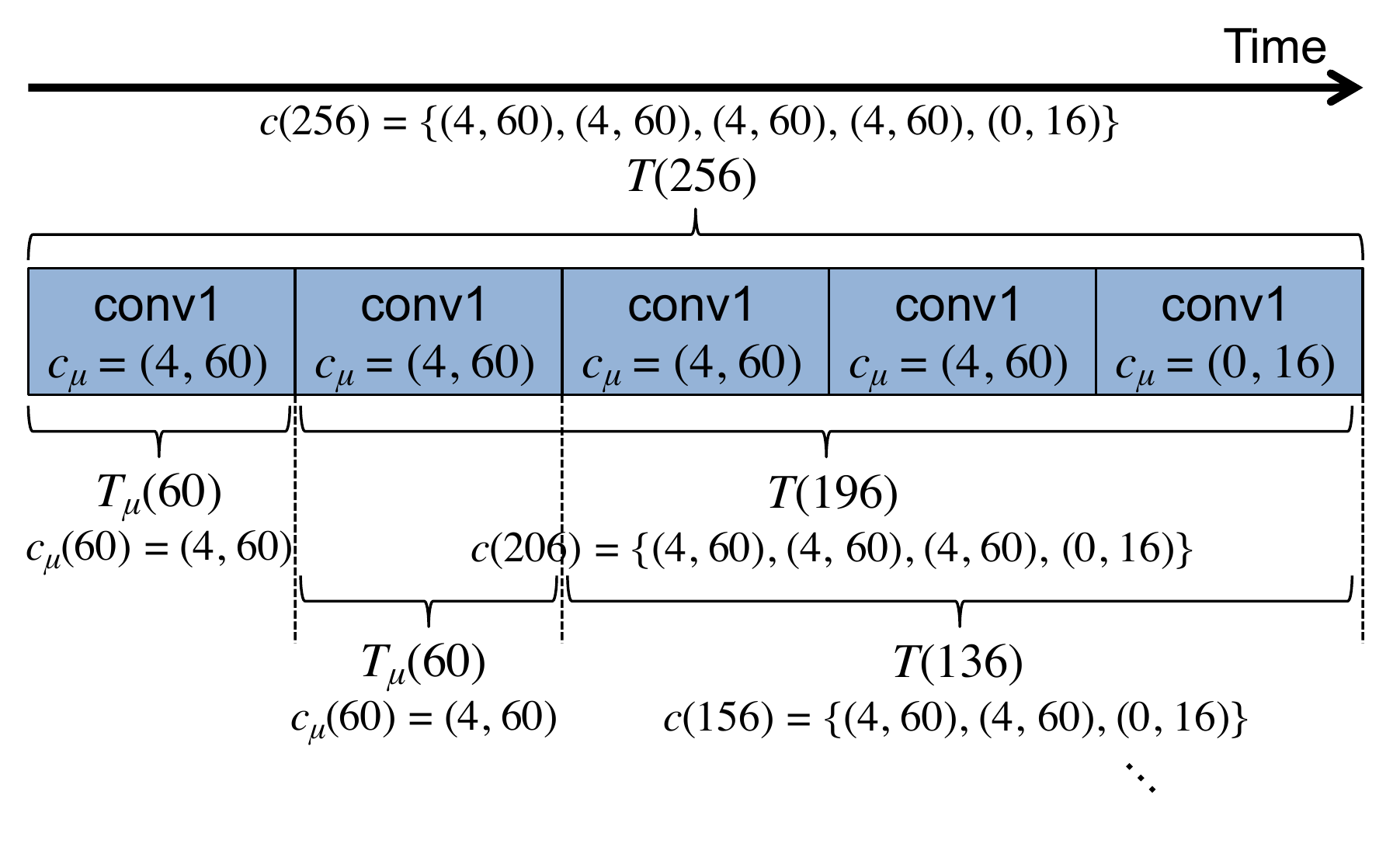}
  \caption{DP-based optimization of \wreuse.
    Here we assume that convolution algorithm 4 with micro-batch size of 60 ($c_\mu(60) = (4, 60)$) achieves better computation efficiency, hence it is repeatedly used.}
  \label{figure:dp}
\end{figure}

The goal of the \wreuse \ policy is to optimize $T(B)$, the total execution time with mini-batch size of $B$ using Dynamic Programming (DP), given by:
\begin{eqnarray}
  T(b) = \min
  \left\{
  \begin{array}{l}
    T_\mu(b), \\
    \min_{b'=1,2,\ldots,B-1}{T(b') + T(b-b')} \\
  \end{array}
  \right\}, \nonumber
\end{eqnarray}
where $T_\mu(b)$ is the fastest execution time of one convolution kernel with a micro-batch size of $b$, within the workspace constraint.
If the first row of the definition of $T(B)$ is smaller than the second row, \ucudnn \ does not have to divide the batch.
Otherwise, it is beneficial to divide the batch into two or more parts, applying the process recursively (\figref{figure:dp}).

The key point of \wreuse \ is that the optimal micro-configuration size is deterministic and independent from other kernels.
This is because in this case, we assume that multiple kernels do not run simultaneously.

The algorithm of \wreuse \ is three-fold, where the mini-batch size is $B$, and user-given maximum workspace size is $M$:
\begin{enumerate}
\item For $b=1, 2, \cdots, B$, \wreuse \ benchmarks all available convolution algorithms of micro-batch size of $b$ with maximum workspace size of $M$, using \cudnn. We define the fastest micro-configuration as $c_\mu(b) = (a, b)$ (where $a$ is the fastest algorithm) and its execution time as $T_\mu(b)$. \label{algorithm:wr__benchmark_kernel}
\item For $b=1, 2, \cdots, B$, \wreuse \ computes $T(b)$, the fastest execution time for micro-batch size of $b$, and $c(b)$, the corresponding configuration, as follows (where $T(0) = 0, c(0) = \emptyset$). $T(b)$ and $c(b)$ are memorized and reused for further iterations.
  \begin{eqnarray}
    \hat{b_\mu} &\leftarrow& \argmin_{b_\mu = 1,2,\ldots,b}\{ T_\mu(b_\mu) + T(b-b_\mu) \} \nonumber \\
    T(b) &\leftarrow& T_\mu(\hat{b_\mu}) + T(b-\hat{b_\mu}) \nonumber \\
    c(b) &\leftarrow& \{c_\mu(\hat{b_\mu})\} + c(b-\hat{b_\mu}) \nonumber
  \end{eqnarray}
\item Outputs the optimal configuration $c(B)$.
\end{enumerate}

\subsection{\wdiv \ Algorithm}
In the \wdiv \ scheme, configurations for multiple convolution kernels are optimized, while at the same time the total workspace size should be less than the total workspace limit that users specify.
Therefore, \wdiv \ is a more complex problem than \wreuse, since the configuration of each convolution kernel is no longer independent from others, due to the total workspace size constraint.

\begin{figure}[t]
  \centering
  \includegraphics[width=\figwidth]{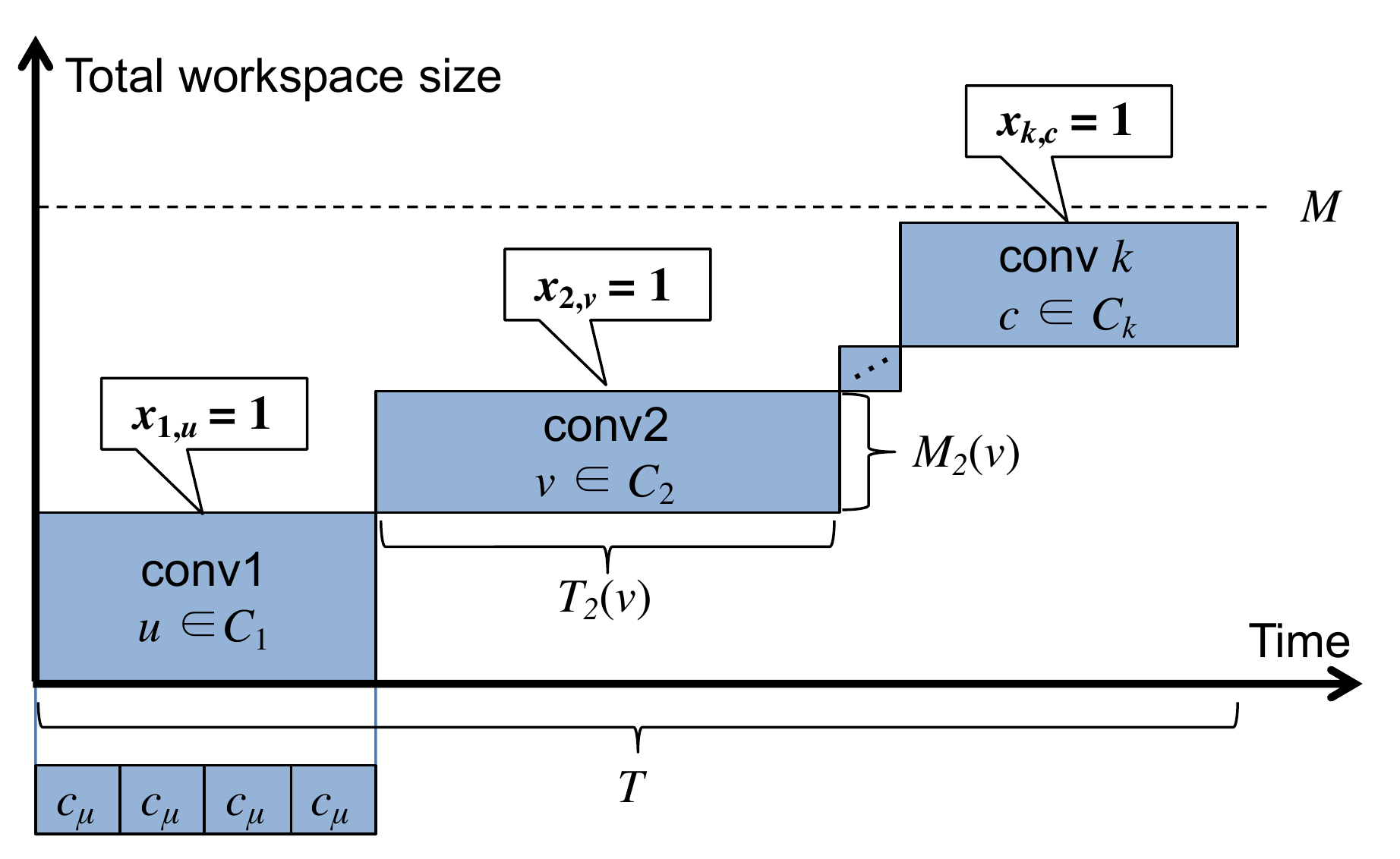}
  \caption{ILP-based optimization of \wdiv.
    The problem is stacking ``time $\times$ memory'' rectangles of configurations diagonally,
    and obtaining the minimum total width $T$,
    provided that the total height is lower than $M$.
    Each configuration $u,v,\ldots,c$ is composed of one or more micro-configurations such as $c_\mu$.}
  \label{figure:ilp}
\end{figure}

To solve this problem, we formulate a 0-1 Integer Linear Programming (ILP)-based optimization algorithm (\figref{figure:ilp}).
Given the set of kernels $\kernelset$ and sets of available configurations $C_k$ of kernel $k$, \wdiv \ is solved by minimizing Equation \ref{eqn:wd}:
\begin{eqnarray}
  {\rm min.} && T = \sum_{k \in \kernelset} \sum_{c \in C_k} T_k(c) x_{k,c} \label{eqn:wd} \\
  {\rm subject\ to.} && \sum_{k \in \kernelset} \sum_{c \in C_k} M_k(c) x_{k,c} \leq M \label{eqn:wd_memory} \\
             && \sum_{c \in C_k} x_{k,c} = 1 \ (\forall k \in \kernelset) \label{eqn:wd_config_per_kernel} \\
             && x_{k,c} \in \{0, 1\} \ (\forall k \in \kernelset, \forall c \in C_k) \label{eqn:wd_end} ,
\end{eqnarray}
where $M_k(c)$ and $T_k(c)$ are the workspace size and execution time of kernel $k$ with configuration $c$, respectively.
Equation \ref{eqn:wd_memory} limits the total workspace size to the user-specified size $M$.
\ucudnn \ uses configuration $c$ on kernel $k$ if and only if $x_{k,c} = 1$, and exactly one of them is selected for each kernel $k$, according to the constraint in Equation \ref{eqn:wd_config_per_kernel}.

\subsubsection{Desirable Configuration Selection}
The challenging problem of the above ILP-based algorithm is that if all possible configurations are evaluated (i.e., all combinations of the number of micro-batch and algorithms),
the search-space is in the order of $|\kernelset| |A|^B$ (where $A$ is set of algorithms and $B$ is the mini-batch size) configurations in total, which makes the problem impractically large.

Here we compute a Pareto front to remove undesirable configurations from all possible configurations, without returning any sub-optimal solutions.
The resulting Pareto front $C_k$ is then input to the ILP (Equation \ref{eqn:wd}-\ref{eqn:wd_end}) to solve the entire problem.

First, we modify the DP algorithm from WR (Section \ref{subsection:wr_algorithm}) to output a set of configurations, rather than the fastest configuration, as follows:
{\small
  \begin{eqnarray}
    C(b)=
    D \Bigg(
    \bigcup_{b_\mu=1,2,\ldots,b}
    \ \bigcup_{c_\mu \in C_\mu(b_\mu)}
    \ \bigcup_{c \in C(b-b_\mu)}
    ( \{ c_\mu \} + c )
    \Bigg), \nonumber
  \end{eqnarray}
}where $C_\mu(b)$ is a set of available micro-configurations of micro-batch size of $b$, and $D$ is a pruning function described below.
Note that this outputs $c(B)$ of the \wreuse \ algorithm as one of its elements;
$c(b) \in C(b)$ and $c_\mu(b) \in C_\mu(b)$ for any $b$.

\begin{figure}[t]
  \centering
  \includegraphics[width=\smallfigwidth]{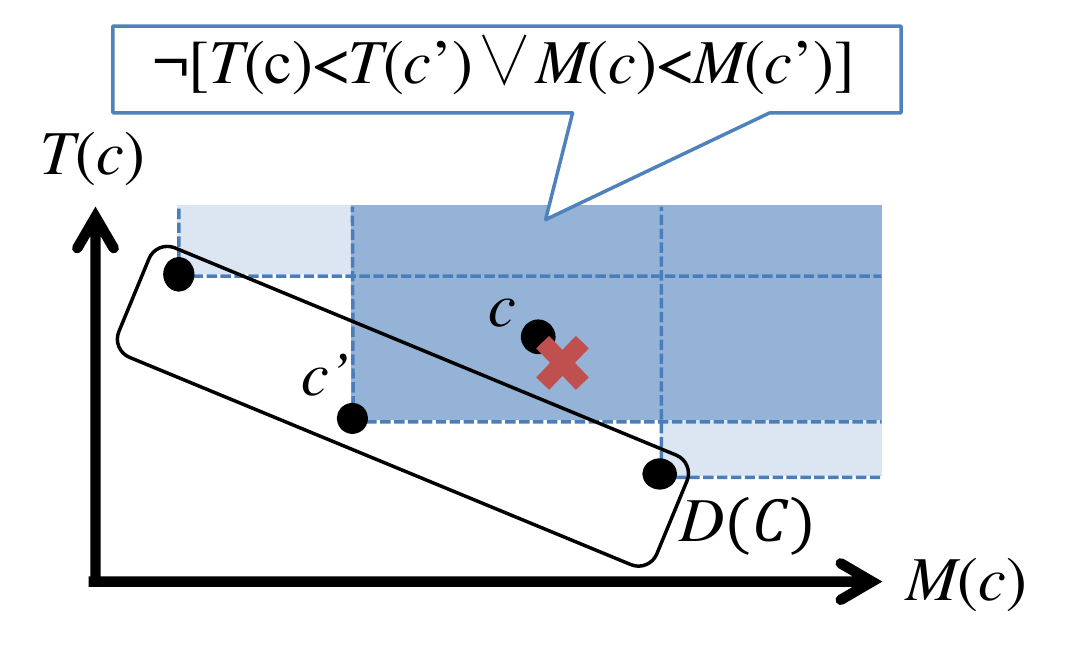}
  \caption{The concept of desirable set. Here $c$ cannot be in $D(C)$ because a $c'$ exists for which the condition $T(c) < T(c') \lor M(c) < M(c')$ is not satisfied.}
  \label{figure:desirable_set}
\end{figure}

Second, we define the ``desirable configuration set'' $D(C) \subset C$ as a Pareto front in the two-dimensional (execution time $\times$ workspace size) space (\figref{figure:desirable_set}):
{\small
  \begin{eqnarray}
    D(C)=\{c \in C | \forall c' \in C \ [ T(c) < T(c') \lor M(c) < M(c') ] \}, \nonumber
  \end{eqnarray}
}where $T(c)$ and $M(c)$ is execution time and required workspace size of a configuration set $c$.
This definition implies that any $c \in D(C)$ is the fastest configuration among any of the elements of $D(C)$ using a workspace size of $M(c)$ or less.
Conversely, if an element $c \in C$ is not in $D(C)$, there is an element that is faster than $c$ and requires less workspace, hence there is no reason to choose $c$, namely ``undesirable''.

\begin{figure}[t]
  \centering
  \includegraphics[width=\figwidth]{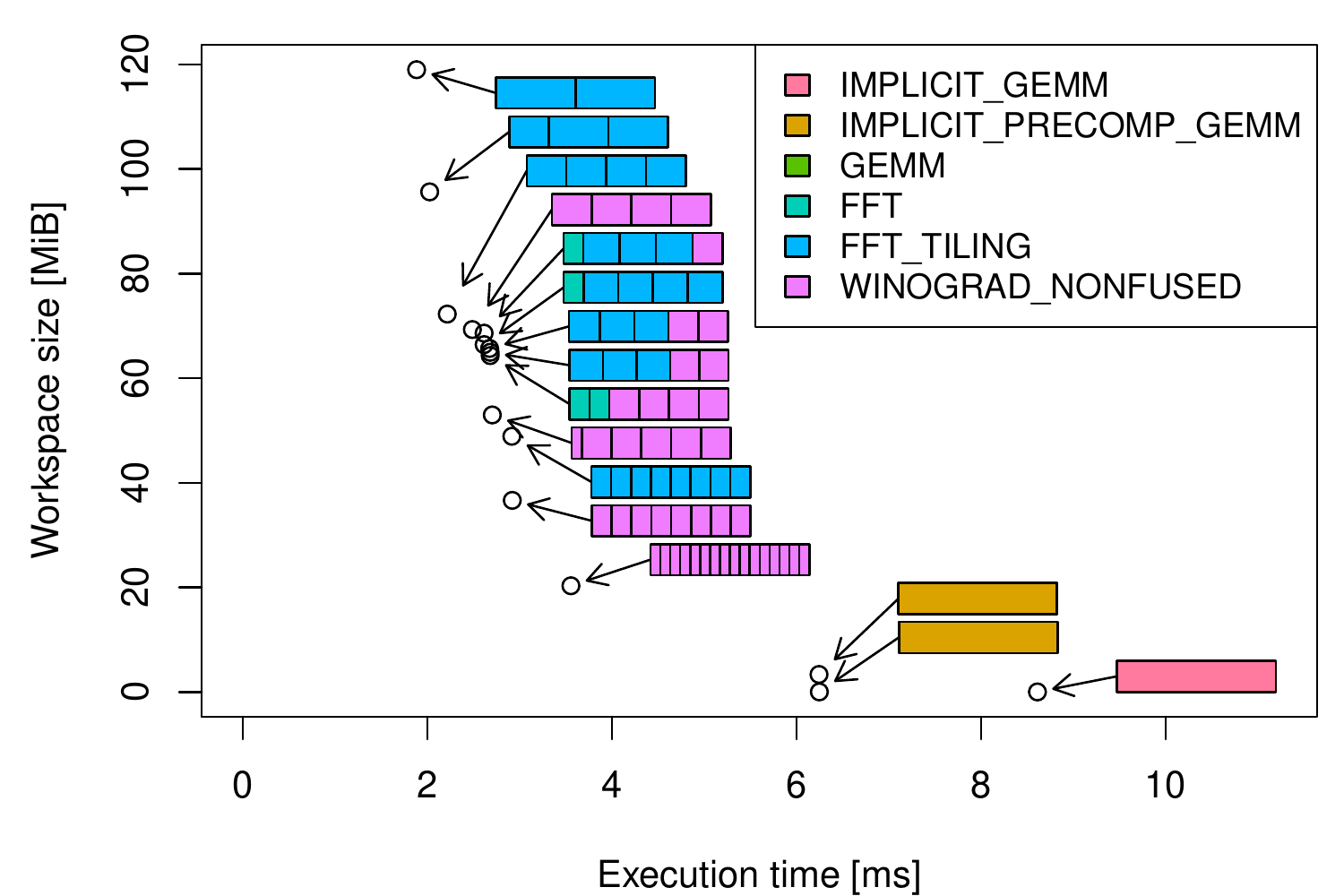}
  \caption{Desirable configurations (i.e. a Pareto front) of \alexnet's ``conv2'' layer (Forward) on \phundred \ with a maximum workspace size of 120 MiB, and a mini-batch size of 256.
    Colored bars corresponding to data points represent the division of the mini-batch and the chosen micro-batch algorithms.
    For example, the top-left point divides the mini-batch into two micro-batches of 128 and utilizes the {\tt FFT\_TILING} algorithm.}
  \label{figure:caffe_time_alexnet_p100_smx2_tsubame3_wdiv_desirable_set_4}
\end{figure}

The pruning drastically reduces the number of variables of Equation \ref{eqn:wd}, and enables solving the ILP for state-of-the-art deep CNNs in practical time.
For instance, the maximum number of desirable configurations of \alexnet's layers we examined in Section \ref{subsection:evaluation_wdiv} was 68, which is much smaller than the exponential order.
\figref{figure:caffe_time_alexnet_p100_smx2_tsubame3_wdiv_desirable_set_4} illustrates a Pareto front of one convolutional layer of \alexnet.

The validity of the pruning algorithm that the optimal solution of the ILP does not include any undesirable configurations is proved as follows:
\begin{proof}
  Suppose that an optimal solution of the ILP $f: X \rightarrow \{0,1\}$, where $X$ is the set of variable symbols of the ILP,
  contains an undesirable configuration $u$ of a kernel $a$ (i.e. $f(x_{a,u}) = 1$).
  According to the definition of desirable sets, there is a configuration $v$ of $a$ such that
  $T_{a}(v) \leq T_{a}(u)$ and $M_{a}(v) \leq M_{a}(u)$.
  According to Equation \ref{eqn:wd_config_per_kernel}, $f(x_{a,c}) = 1-f(x_{a,u}) = 0$ for all $c \in C_a$.

  Let $g: X \rightarrow \{0,1\}$ be defined as
  \begin{eqnarray}
    g(x_{k,c}) = \left\{
    \begin{array}{ll}
      1 & (k = a \land c = v) \\
      0 & (k = a \land c \neq v) \\
      f(x_{k,c}) & (\mathrm{otherwise})
    \end{array}
                   \right.
                   \nonumber .
  \end{eqnarray}
  $g$ satisfies Equation \ref{eqn:wd_config_per_kernel} for $k = a$ as
  \begin{eqnarray}
    \sum_{c \in C_a} g(x_{a,c}) &=& \sum_{c \in C_a \backslash \{v\}} g(x_{a,c}) + g(x_{a,v}) = 1 \nonumber ,
  \end{eqnarray}
  and Equation \ref{eqn:wd_memory} as
  \begin{eqnarray}
    \sum_{k \in \kernelset} \sum_{c \in C_k} M_k(c) g(x_{k,c})
    &=& \sum_{k \in \kernelset \backslash \{a\}} \sum_{c \in C_k} M_k(c) g(x_{k,c}) \nonumber \\
    && + M_a(v) g(x_{a,v}) \nonumber \\
    &\leq& \sum_{k \in \kernelset \backslash \{a\}} \sum_{c \in C_k} M_k(c) f(x_{k,c}) \nonumber \\
    && + M_a(u) f(x_{a,u}) \nonumber \\
    &\leq& \sum_{k \in \kernelset} \sum_{c \in C_k} M_k(c) f(x_{k,c}) \leq M \nonumber .
  \end{eqnarray}
  Similarly, by replacing $M_k$ as $T_k$ in the inequality above, the objective value of $g$ is proved to be lower than $f$,
  hence $g$ is a better solution of the ILP.
  Therefore it contradicts the supposition that $f$ is the optimal solution.
\end{proof}

\subsection{\ucudnn \ Implementation} \label{subsection:ucudnn_implementation}
To enable \ucudnn, the only modification that needs to be performed to the code is to replace the \cudnn \ handle type \handle \ with \uhandle.
The \ucudnn \ handle object is an opaque type that wraps the original type, such that users can call any \cudnn \ function.
When a convolution operation or benchmarking function is called with the \ucudnn \ handle object, the \ucudnn \ library internally computes the optimal configurations, and returns a virtual algorithm ID and zero required workspace size.
This mechanism enables users to call \ucudnn \ with minimal modification to the original code.
For example, the number of lines to be modified to introduce \ucudnn \ to \caffe \ (v1.0) is approximately three.

The implementation of \ucudnn \ is based on overloading a subset of \cudnn \ functions,
where the memory of the \ucudnn \ handle type is structured to behave to act as the \cudnn \ internal handle for the other calls.
We define a cast operator from the \ucudnn \ handle to \cudnn \ handle so that the framework automatically adopts this method.
Using this technique, \ucudnn \ delegates most of the functions to \cudnn, but overrides functions related to the convolutional layers.

The optimization algorithm in \ucudnn \ is based on the methodology described in Section \ref{subsection:ucudnn_methodology}.
In practice, \ucudnn \ provides a ``batch size policy'', which determines what micro-batch sizes are benchmarked at the step \ref{algorithm:wr__benchmark_kernel} of the \wreuse \ algorithm, as follows:
\begin{itemize}
\item \ttall \ uses all batch sizes $b \in \{1, 2, 3, \cdots, B\}$. Although this always finds the optimal solution, it takes $\mathcal{O}(B)$ time for the benchmark.
\item \ttpoweroftwo \ uses only power-of-two batch sizes $b \in \{2^0, 2^1, 2^2, \cdots, B\}$. This saves a considerable amount of time since it only costs $\mathcal{O}(\log B)$ time for the benchmark.
\item \ttundivided \ uses only the original mini-batch size $b \in \{B\}$. In \wreuse, this option always selects the same configuration as \cudnn, hence this option is only useful to evaluate the overhead of \ucudnn.
\end{itemize}
These policies can be specified via an environment variable or through a special library function in \ucudnn.
Furthermore, \ucudnn \ supports parallel micro-configuration evaluation via an environment variable, in which the aforementioned micro-batches are distributed to different GPUs on the same computing node and tested concurrently.
This function assumes that the node contains multiple homogeneous GPUs.

\ucudnn \ caches the optimized configurations and the benchmark results into memory and optional file-based database respectively, to skip unnecessary recomputations.
This is especially beneficial for networks that replicate convolutional layers of the same size, such as \resnet \ \cite{He2016}.
In addition, the file-based caching enable offline benchmarking, as well as sharing the results among a homogeneous GPU cluster via network file system.

\subsection{Implementation of \wdiv \ Optimization}
To perform \wdiv \ optimization, \ucudnn \ must know the number of convolutional layers and corresponding layer parameters in advance, i.e., before running any kernel.
In the current \cudnn \ API, however, the parameters are passed one layer at a time, and thus there is no way to obtain all the parameters collectively from deep learning frameworks.

To overcome this issue, we assume that the deep learning framework calls {\tt cudnnGetConvolution*Algorithm} one time for each layer prior to the computation of the entire network (e.g., training, inference).
This is the most straightforward use of the \cudnn \ interface, as memory (including workspace) is usually allocated before initiating computations.
Due to the specific implementation of \caffe, we add a \ucudnn \ library call after network initialization, which ignores subsequent {\tt cudnnGetConvolution*Algorithm} calls.

When {\tt cudnnGetConvolution*Algorithm} is called, \ucudnn \ pushes the kernel parameters to an internal list, and returns a dummy result.
Note that the returned results satisfy the semantics given by the \cudnn \ interface, so the framework will not raise errors and will not allocate its own workspaces.
When {\tt cudnnConvolution*} is called for the first time, \ucudnn \ executes the optimization algorithm (namely, \wdiv).
We use the GNU Linear Programming Kit (GLPK) \cite{glpk} as the ILP solver.

\begin{table}[htbp]
  \centering
  \caption{Evaluation Environment Specification.}
  \label{table:spec}
  \scalebox{0.9}{
    \begin{tabular}{l|c c c}
      \bhline
      & \textbf{\kfc}
      & \textbf{\tthree}
      & \textbf{\dgxone} \\
      \bhline

      \multirow{3}{10mm}{CPU (Intel Xeon)}
      & \multirow{3}{*}{E5-2620 $\times$ 2}
      & \multirow{3}{*}{E5-2680 v4 $\times$ 2}
      & \multirow{3}{*}{E5-2698 v4 $\times$ 2} \\
      &&& \\
      &&& \\
      \hline

      \multirow{4}{10mm}{GPU (NVIDIA Tesla)}
      & \keighty \ $\times$ 4 & \phundred \ $\times$ 4 & \vhundred \ $\times$ 8 \\
      & - 8.73 SP TFlop/s     & - 10.6 SP TFlop/s      & - 15.7 SP TFlop/s      \\
      & - 24 GiB GDDR5        & - 16 GiB HBM2          & - 16 GiB HBM2          \\
      & (480 GiB/s BW) & (732 GiB/s BW)  & (900 GiB/s BW)  \\
      \hline

         &                 & SUSE Linux        &                \\
      OS & CentOS 7.3.1611 & Enterprise Server & Ubuntu 16.04.3 \\
         &                 &  12 SP2           &                \\
      \hline

      CUDA   & 8.0.61          & 8.0.44 & 9.0 \\
      \cudnn & 6.0             & 6.0 & 7.0.5 \\
      GLPK   & 4.63            & 4.63 & N/A \\
      \hline

      \multirow{2}{*}{\caffe}
      & \multirow{2}{*}{1.0} & \multirow{2}{*}{1.0} & NVCaffe \\
      &&& v0.16.5 \cite{nvcaffe} \\
      \hline
      \tensorflow & N/A & 1.4.1 & N/A \\
      \bhline
    \end{tabular}
  }
\end{table}

\section{Performance Evaluation}
We evaluate the performance of \ucudnn \ for three different GPU architectures, NVIDIA Tesla \keighty \ \cite{k80}, \phundred \ \cite{p100} and \vhundred \ \cite{v100} on the \kfc, \tthree, and \dgxone \ supercomputers, respectively.
The specifications of these supercomputers are listed in \tabref{table:spec}.

Throughout the evaluation, we use single-precision floating point format and store tensors in the $NCHW$ storage order.
We use three different deep learning frameworks for evaluations: \caffe \ \cite{jia2014caffe}, its NVIDIA branch (\nvcaffe) \cite{nvcaffe}, and \tensorflow \ \cite{tensorflow2015-whitepaper}.
Both support recent versions of \cudnn \ (6 or 7).
We use a built-in benchmarking command (\caffe's ``time'' command) or an official benchmarking script (from \tensorflow \ models repository \cite{tensorflow_models}) to measure the execution time of forward and backward passes,
and show the sum of forward and backward passes together.
In the following sections, unless explicitly mentioned,
each forward-backward passes are measured 50 times on \caffe \ and 100 times on \tensorflow.

For neural networks, we use \alexnet \cite{Krizhevsky2012}, \resnet \cite{He2016}, and \densenet \cite{huang2017densely}.
For evaluations on \caffe, we use the \alexnet \ model defined in \caffe, \resnet-18, and \resnet-50 from \nvcaffe.
We modify data prefetching size from 4 to 16 for \alexnet \ and \resnet-18 for \tthree.
For evaluations on \tensorflow, we use the definitions in an official benchmarking repository \cite{tensorflow_benchmarks}.


As for workspace limit, unless explicitly mentioned, we use 8 MiB and 64 MiB for each layer, which are the default workspace size limits of \caffe \ and \caffetwo \ \cite{caffe2} \ respectively.
In addition, we use 512 MiB of workspace per layer to investigate the case where sufficiently large workspace is provided.
To shorten the benchmarking time, we use several GPUs on the same node with the parallel evaluation function of \ucudnn, mentioned in Section \ref{subsection:ucudnn_implementation}.

\begin{figure}[b]
  \centering
  \includegraphics[width=\figwidth]{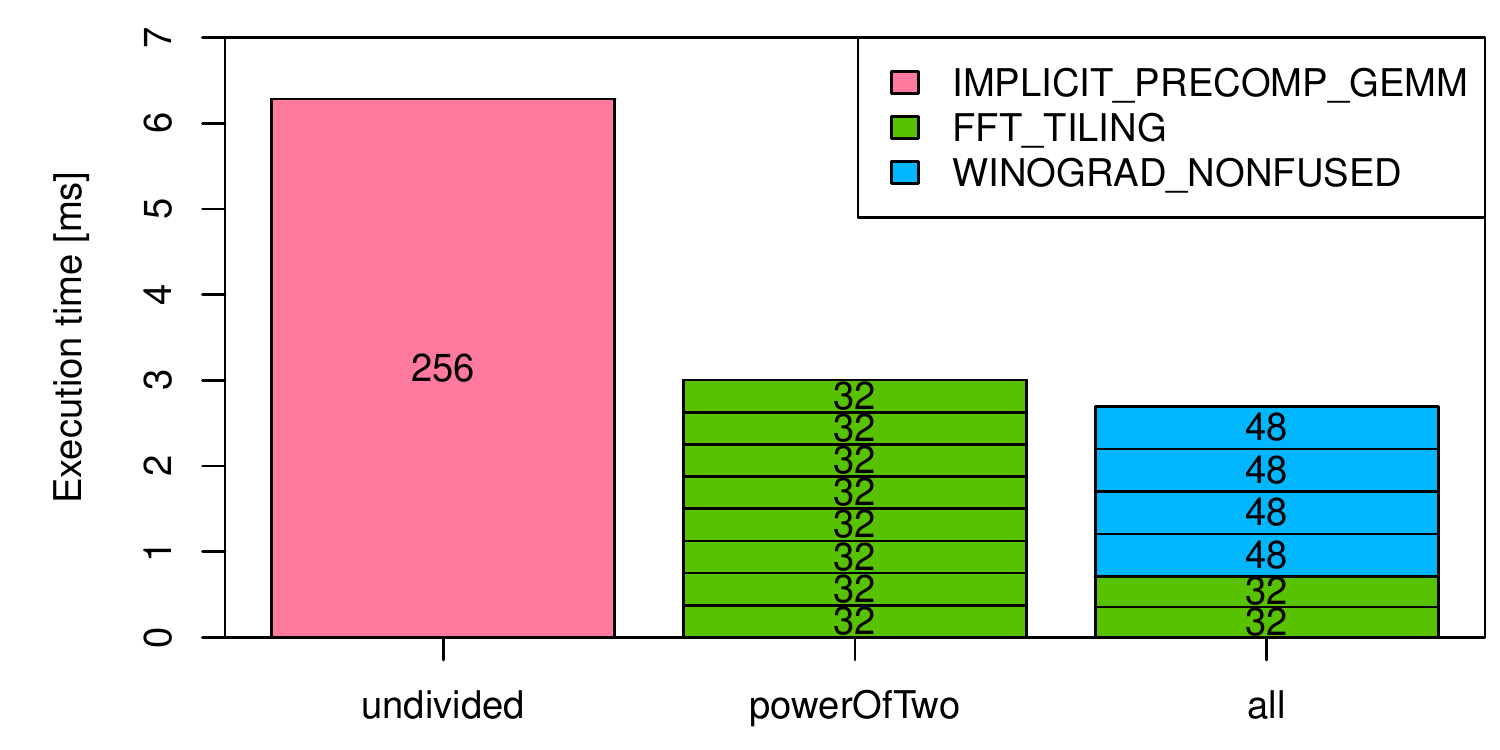}
  \caption{Benchmark results of forward convolution of \alexnet's ``conv2'' layer on \phundred.
    We use 64 MiB workspace size and a mini-batch size of 256.
    Numbers on each rectangle represent micro-batch sizes.}
  \label{figure:caffe_time_alexnet_conv2_forward}
\end{figure}

\begin{figure*}[t]
  \centering
  \subfloat[\keighty]{
    \includegraphics[width=0.3\linewidth]{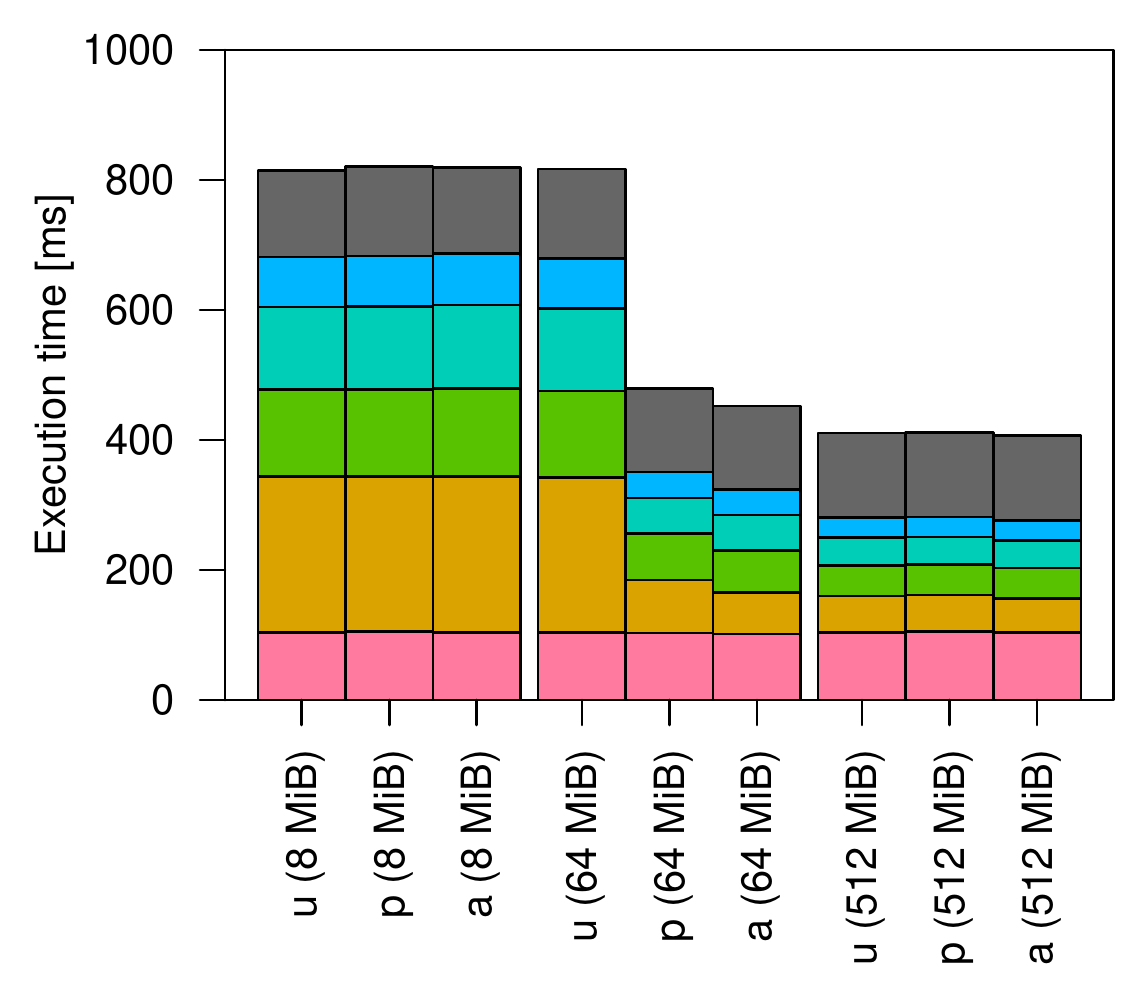}
  }
  \subfloat[\phundred]{
    \includegraphics[width=0.3\linewidth]{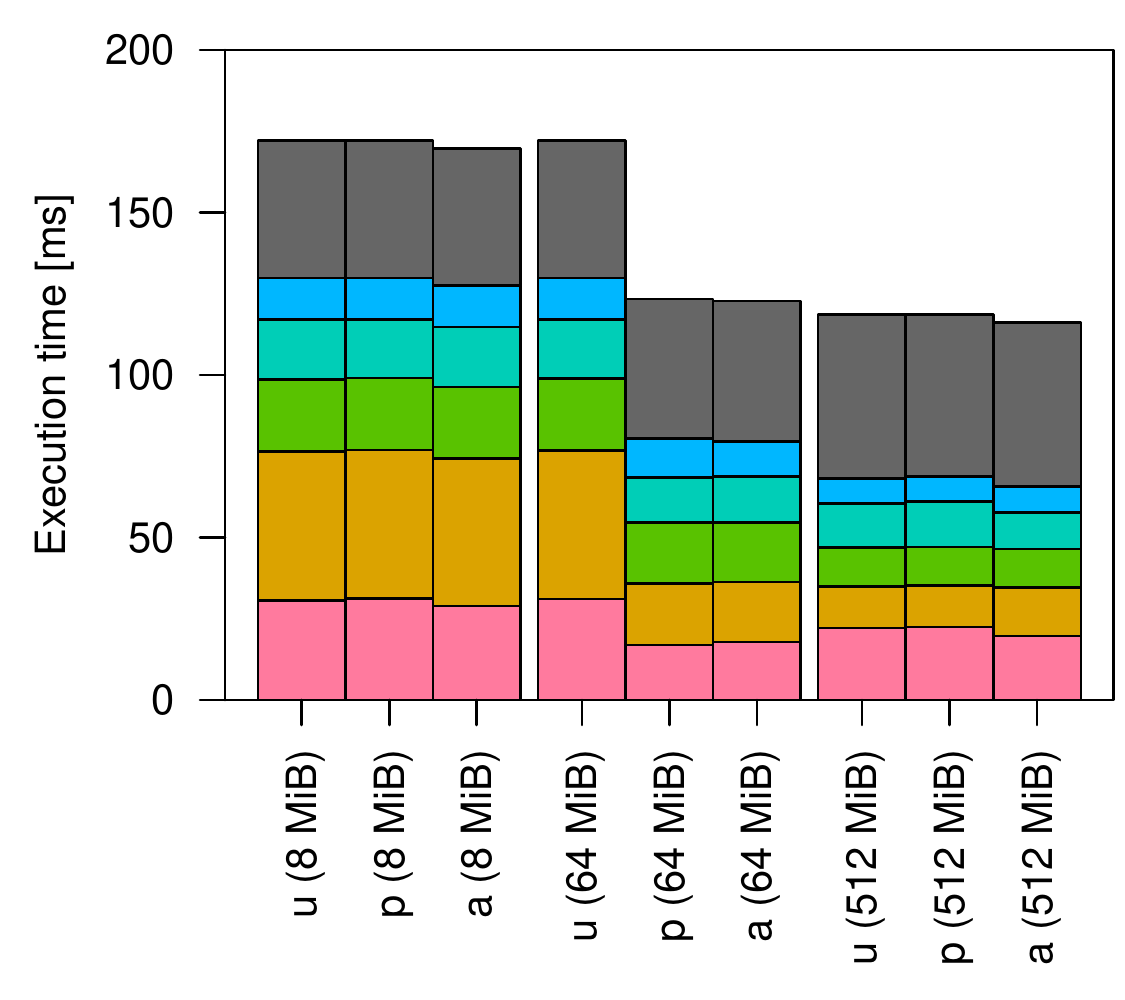}
  }
  \subfloat[\vhundred]{
    \includegraphics[width=0.4\linewidth]{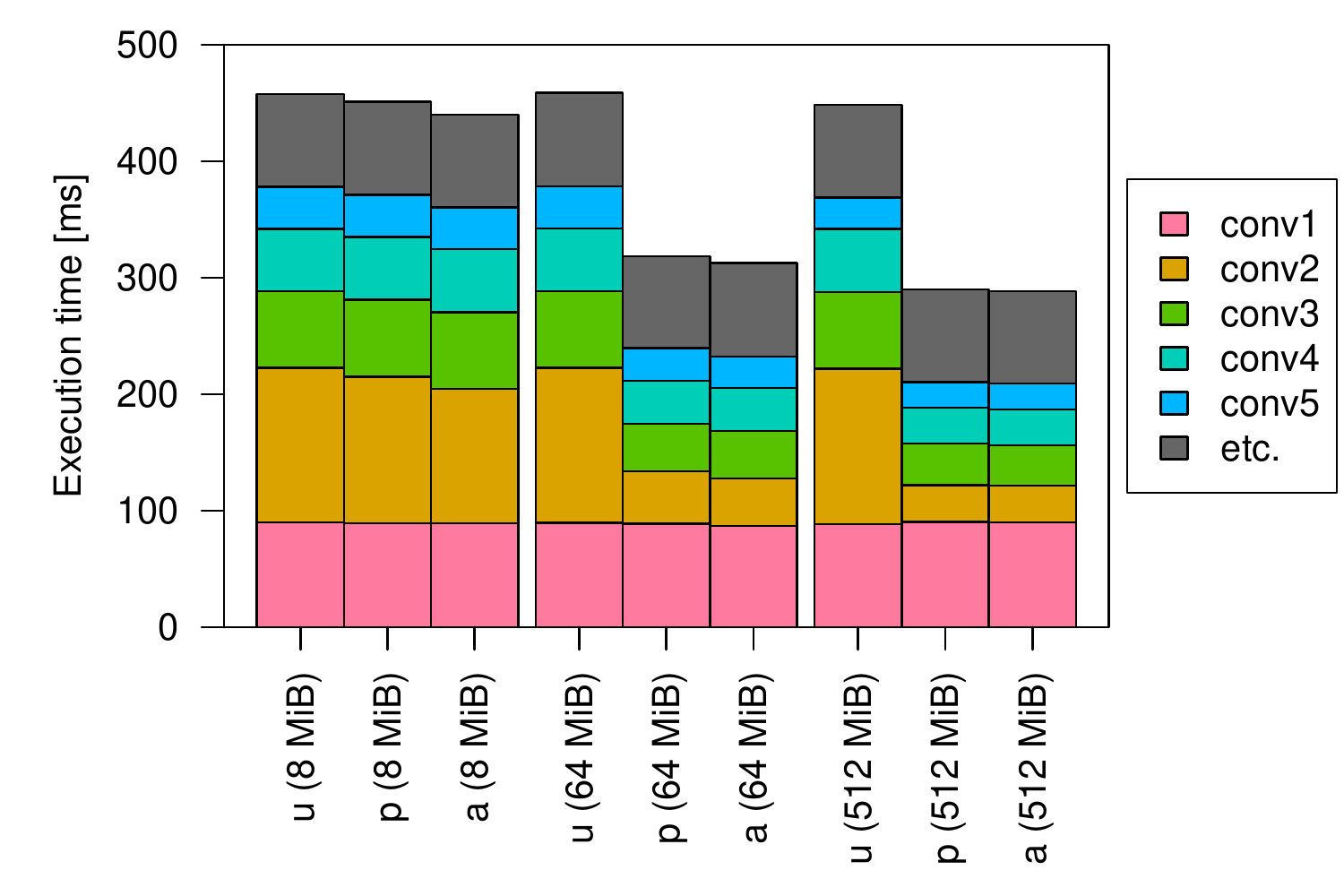}
  }
  \caption{Benchmark results of \alexnet \  on three different GPUs with different workspace sizes (8, 64, 512 MiB).
    The labels ``u'', ``p'' and ``a'' represent \ttundivided, \ttpoweroftwo, and \ttall, respectively.
    We use a mini-batch size of 256 on \keighty \ and \phundred, and 1024 on \vhundred.}
  \label{figure:caffe_time_alexnet}
\end{figure*}

\subsection{Convolution Kernel Optimization Using \wreuse} \label{subsection:evaluation_one_layer}
\figref{figure:caffe_time_alexnet_conv2_forward} shows the execution time of forward convolution ({\tt cudnnConvolutionForward}) of the ``conv2'' layer in \alexnet \ on \phundred.
With workspace size of 64 MiB, the GEMM (GEneral Matrix-Matrix multiply)-based algorithm is the one chosen by \cudnn, requiring only \perf{caffe-time-layer-alexnet-p100-64mb-conv2-forward-gemm-ws} for workspace if the mini-batch is not divided.
On the other hand, FFT-based convolution \cite{b70224ae04784e15b91d2056c46924a6} \ is more efficient, although it requires excessive amount of workspace (\perf{caffe-time-layer-alexnet-p100-64mb-conv2-forward-fft-ws}) to store the images and filters in the frequency domain.
\ucudnn \ with \ttpoweroftwo \ option successfully enables the use of FFT within the workspace size constraints, using \perf{caffe-time-layer-alexnet-p100-64mb-conv2-forward-fft-ws-u32} over micro-batches of size 32.

The \ttall \ option also enables \ucudnn \ to use Winograd convolution \cite{Lavin2016FastAF}, an algorithm that is especially efficient for small convolution kernels, achieving \perf{caffe-time-layer-alexnet-p100-64mb-conv2-forward-speedup}x speedup over \ttundivided \ in total.

\subsection{CNN Optimization Using \wreuse}
We evaluate \wreuse-based optimization on two different deep learning frameworks: \caffe \ and \tensorflow.

\subsubsection{\caffe}

\figref{figure:caffe_time_alexnet} shows timing breakdowns of \caffe \ on \alexnet \  with three different GPUs.
Additionally, we only highlight convolutional layers since the others (e.g., pooling) are out of the scope of this paper.

One important observation from \figref{figure:caffe_time_alexnet} is that the performance improvement of \ucudnn \ over \cudnn \ (which is equivalent to \ttundivided) is significant when the moderate amount of workspace is set by users.
For instance, if the workspace size per kernel is 64 MiB, \ucudnn \ with the \ttall \ option achieves \perf{caffe-time-alexnet-k80-64mb-speedup}x speedup with respect to the entire iteration,
and \perf{caffe-time-alexnet-k80-64mb-speedup-conv}x with respect to convolutions alone, than \ttundivided \ on \keighty.
This is because \ucudnn \ successfully enables \cudnn \ to use faster algorithms, as in the example from Section \ref{subsection:evaluation_one_layer}.
In addition, a similar speedup is achieved on \phundred \ (\perf{caffe-time-alexnet-p100-64mb-speedup}x for the entire iteration, and \perf{caffe-time-alexnet-p100-64mb-speedup-conv}x for convolutions alone),
and on \vhundred \ (\perf{caffe-time-alexnet-v100-64mb-speedup}x for the entire iteration, and \perf{caffe-time-alexnet-v100-64mb-speedup-conv}x for convolutions alone).

In the case where workspace size is limited to 8 MiB, \ucudnn \ cannot attain any performance improvement, because even if the mini-batch is finely divided, the specified workspace is too small to utilize.
Indeed, on \phundred, only one kernel of \ttall \ option seems to increase the utilization of the workspace over \ttundivided.

On the other hand, when the workspace size limit is too large (512 MiB) on \keighty \ and \phundred \ GPUs, performance difference between \cudnn \ and \ucudnn \ is negligible.
This is because there is no benefit from dividing the mini-batch, as all algorithms fit into the workspace constraints.
However, this workspace limit consumes a considerable amount of workspace memory:
While the \ttundivided \ option consumes \perf{caffe-time-alexnet-p100-512mb-ws-undivided} in total, \ttall \ with 64 MiB limit only consumes \perf{caffe-time-alexnet-p100-64mb-ws-all},
although with \perf{caffe-time-alexnet-p100-64mb-speedup-512mb-undivided-percent} overhead caused by the choice of micro-batch algorithms.

From the viewpoint of the time to optimization, including kernel benchmarking and solving DP, \ttpoweroftwo \ considerably outperforms \ttall.
In particular, with 64 MiB workspace on \phundred, \ttall \ takes \perf{caffe-time-alexnet-p100-64mb-benchmark-all}, whereas \ttpoweroftwo \ takes \perf{caffe-time-alexnet-p100-64mb-benchmark-poweroftwo}.
This result and \figref{figure:caffe_time_alexnet} imply that \ttpoweroftwo \ is a reasonable choice to test the computation efficiency of new CNNs quickly.
Generally, the overhead of \ucudnn \ is negligible with respect to the entire training time, in which the forward and backward passes are repeated hundreds of thousands of times.



\begin{figure*}[t]
  \centering
  \subfloat[\alexnet]{
    \includegraphics[width=0.28\figwidth]{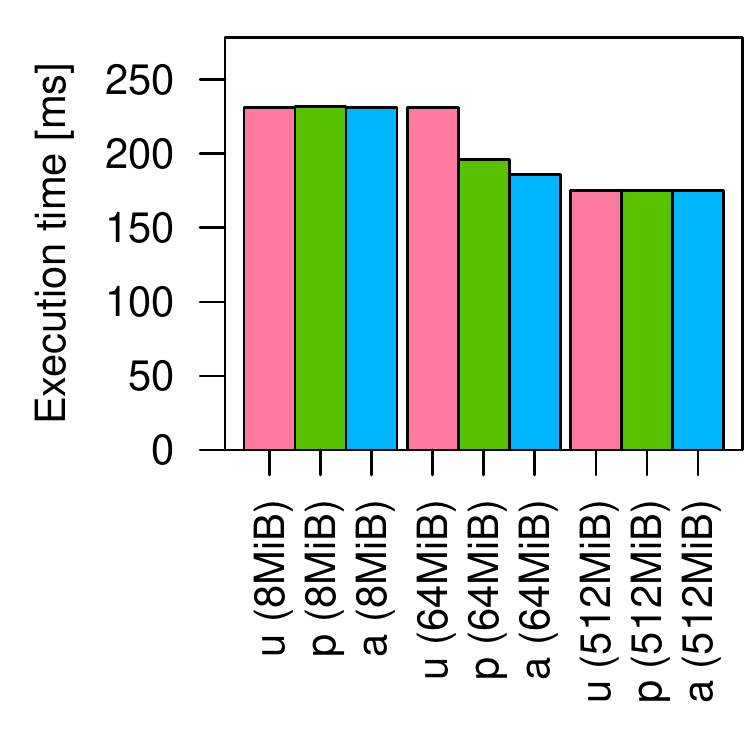}  
  }
  \subfloat[\resnet-50]{
    \includegraphics[width=0.28\figwidth]{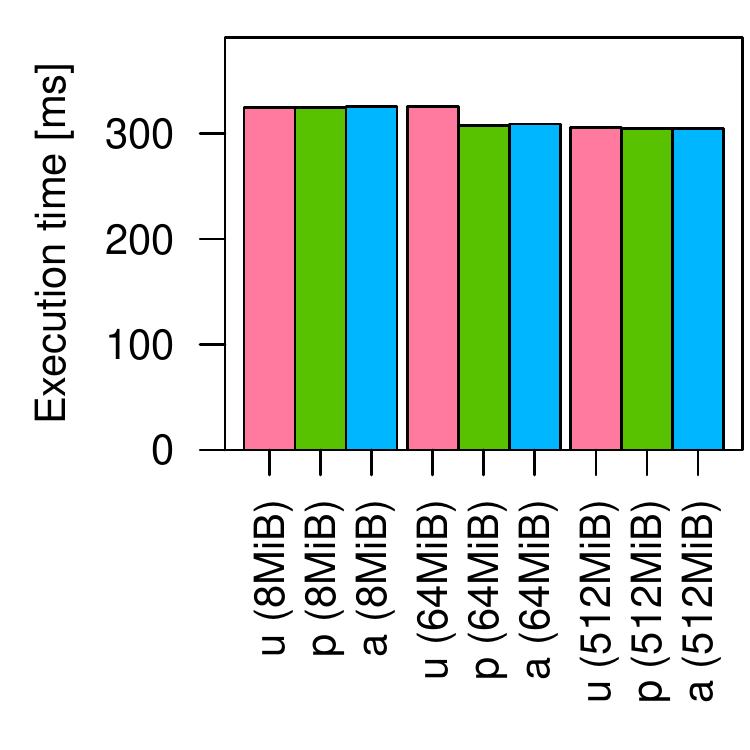} 
  }
  \subfloat[\densenet-40 ($k=40$)]{
    \includegraphics[width=0.42\figwidth]{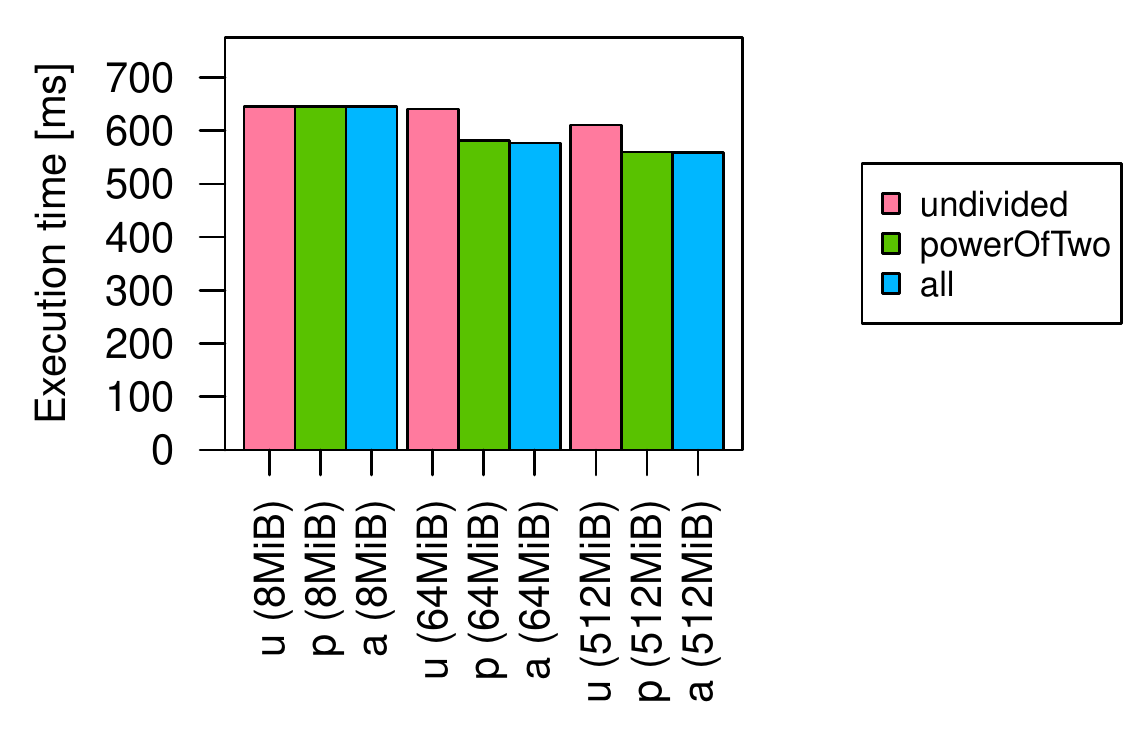} 
  }
  \caption{Benchmark results of different CNNs on \phundred \ with different workspace sizes (8, 64, 512 MiB), using \tensorflow \ framework.
    We use a mini-batch size of 256 for \alexnet \ and \densenet, and 64 for \resnet-50.}
  \label{figure:tensorflow_time}
\end{figure*}

\subsubsection{\tensorflow}
\figref{figure:tensorflow_time} presents timing breakdowns of \alexnet \ and \resnet-50, \densenet-40 on \phundred.

We set the (input width, output width) as $(224, 1000)$ for \alexnet \ and \resnet-50, or $(32, 10)$ for \densenet-40, which are used for training ILSVRC2012 classification dataset \cite{ilsvrc} or the CIFAR dataset \cite{cifar}, respectively.
We also set $k$ of \densenet-40, the number of feature maps of each convolutional layer, to 40 to obtain better computational efficiency.

Since \tensorflow \ 1.4.1 does not provide any workspace limits to \ucudnn \ via \cudnn's benchmarking functions  before actual convolutions,
we manually provide workspace limits of 8, 64, and 512 MiB to \ucudnn.
\ucudnn \ with a workspace limit of 64 MiB achieves \perf{tf-time-alexnet-p100-64mb-speedup}x speedup for \alexnet, and \perf{tf-time-resnet50-p100-64mb-speedup}x for \resnet-50.
These results prove that \ucudnn \ has good performance portability between different deep learning frameworks that depend on \cudnn.

\subsection{Memory Consumption Using \wreuse}
\begin{figure}[!b]
  \centering
  \subfloat[\alexnet \ (\cudnn)] {\includegraphics[width=\halffigwidth]{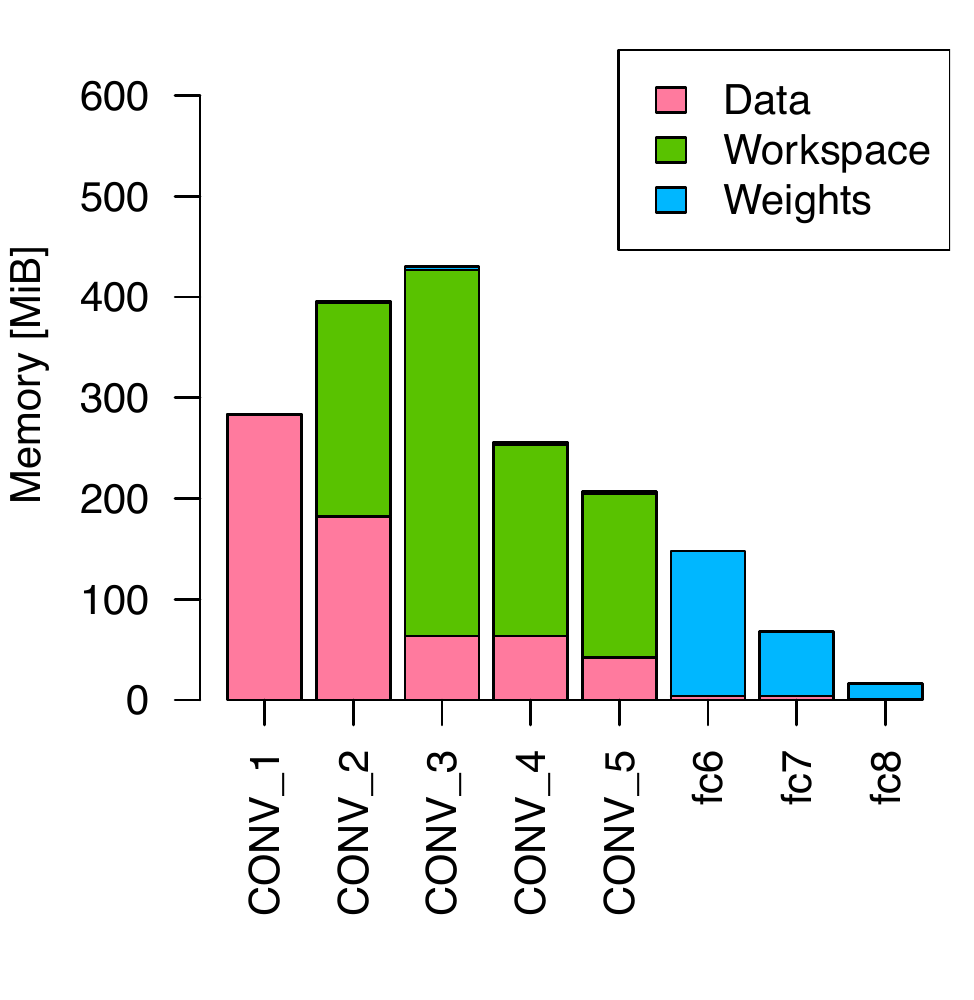}}
  \subfloat[\alexnet \ (\ucudnn)]{\includegraphics[width=\halffigwidth]{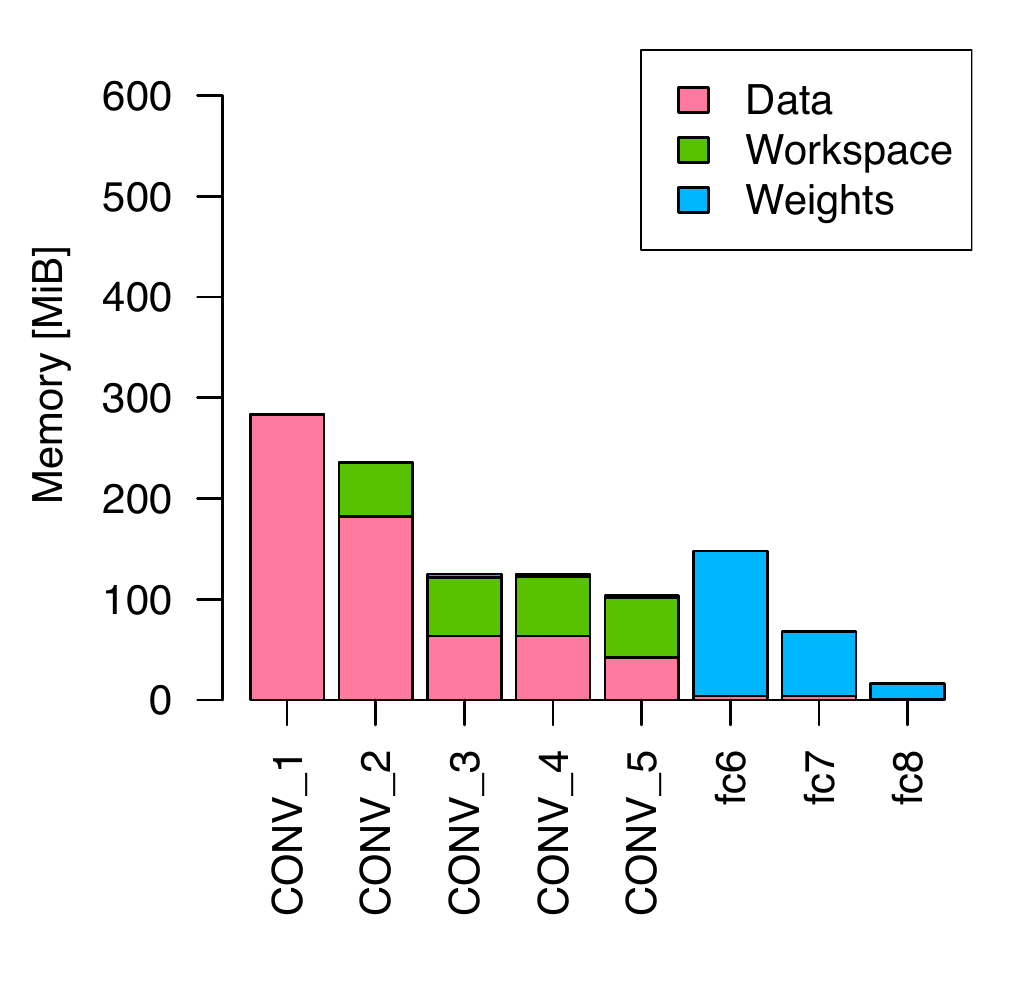}}
  \newline
  \subfloat[\resnet-18 \ (\cudnn)] {\includegraphics[width=\halffigwidth]{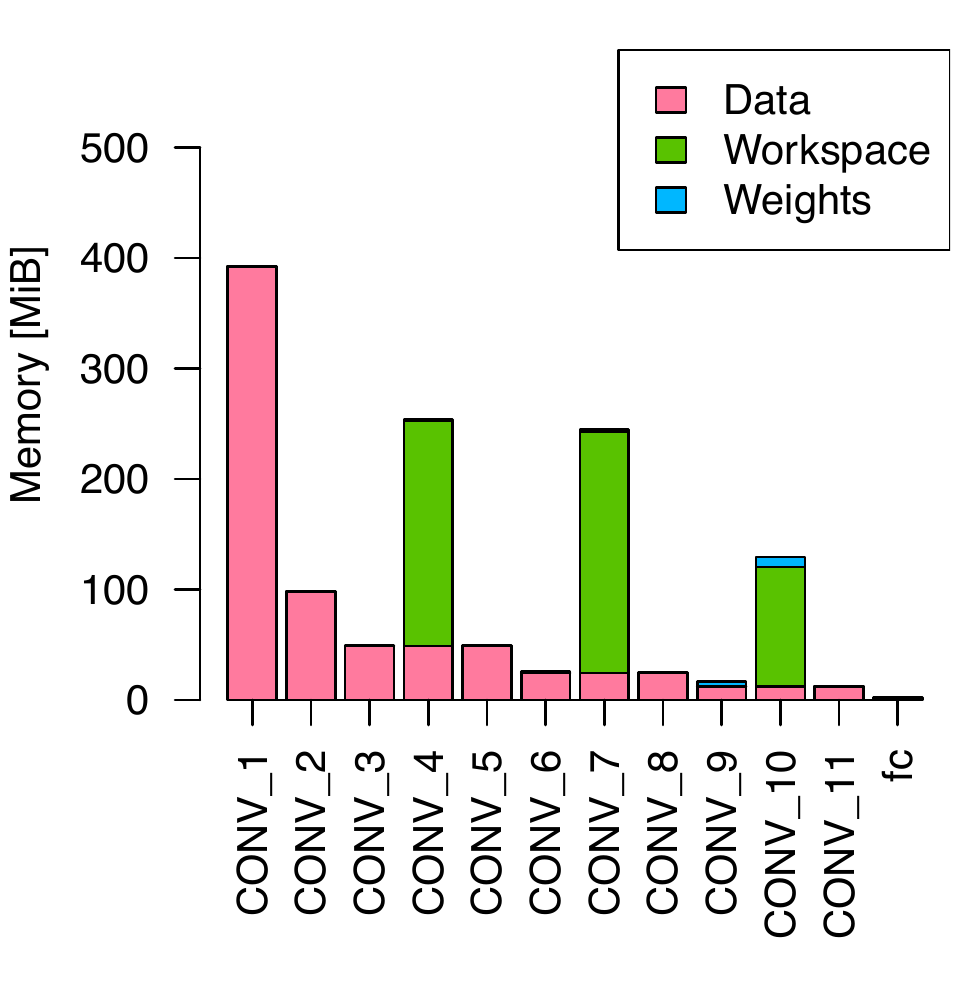}}
  \subfloat[\resnet-18 \ (\ucudnn)]{\includegraphics[width=\halffigwidth]{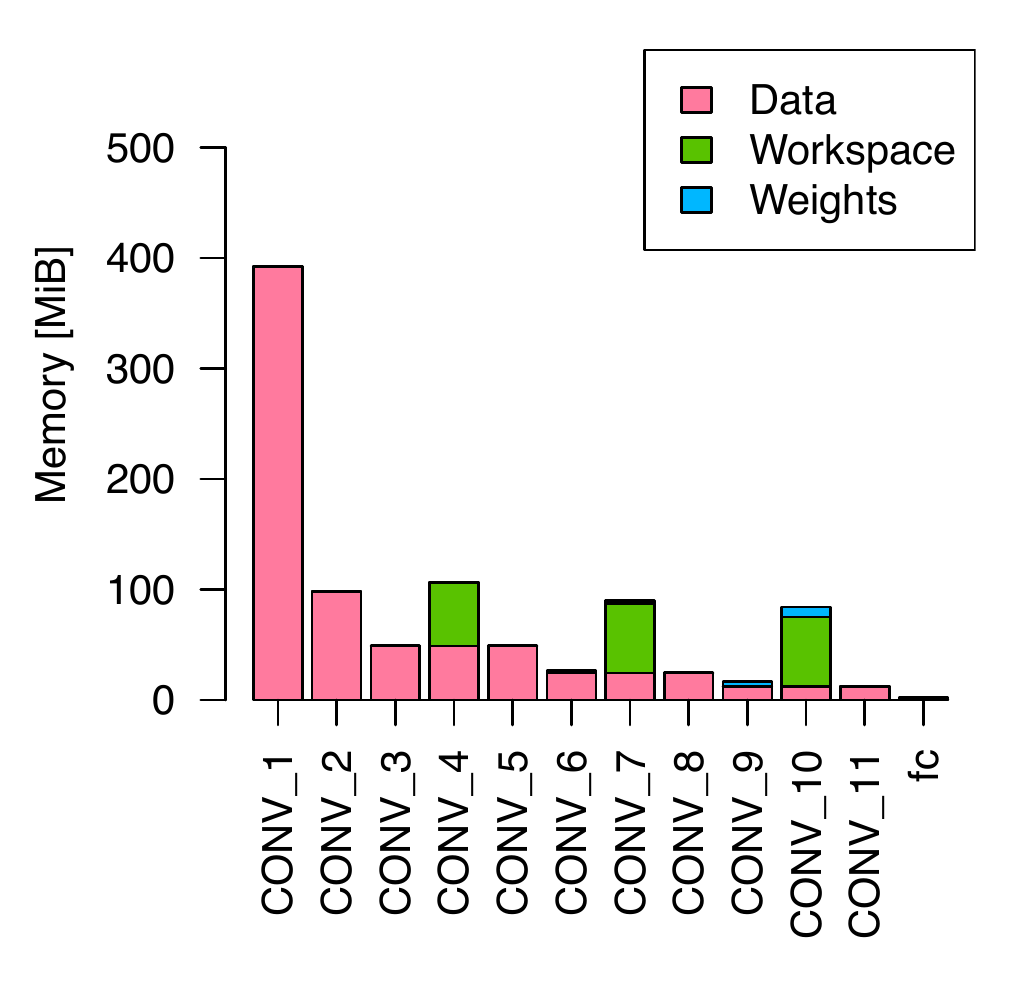}}
  \caption{
    Per-layer breakdowns of memory consumption of \alexnet \ and \resnet-18 on \phundred.
    For simplicity, we only show the memory usage of unique convolutional layers (CONV\_$n$) and fully-connected layers (fc or fc$n$) in one forward propagation.
    We use a mini-batch of 256 for \alexnet \ and 128 for \resnet-18 respectively.
    We set a per-layer workspace limit of 512 MiB for \cudnn, and 64 MiB for \ucudnn.
    Each bar segment of ``WS (\ucudnn)'' represents the maximum workspace size of the layer.
  }
  \label{figure:forward_memory}
\end{figure}

\figref{figure:forward_memory} shows the per-layer memory usage of \alexnet \ and \resnet-18 on \phundred.
In \figref{figure:forward_memory}, we set a per-layer workspace limit of 512 MiB for \cudnn, and 64 MiB for \ucudnn,
where the slowdown due to the decrease of memory limit is negligible (\perf{caffe-time-alexnet-p100-64mb-slowdown-conv-to-512mb-undivided}x).
These figures clearly show that \ucudnn \ can cut down per-layer memory consumption by up to \perf{forward-memory-consumption-ratio-max-alexnet}x and \perf{forward-memory-consumption-ratio-max-resnet18}x on \alexnet \ and \resnet-18 respectively.

\subsection{CNN Optimization Using \wdiv} \label{subsection:evaluation_wdiv}
\begin{figure}[t]
  \centering
  \subfloat[\alexnet]{\includegraphics[width=\figwidth]{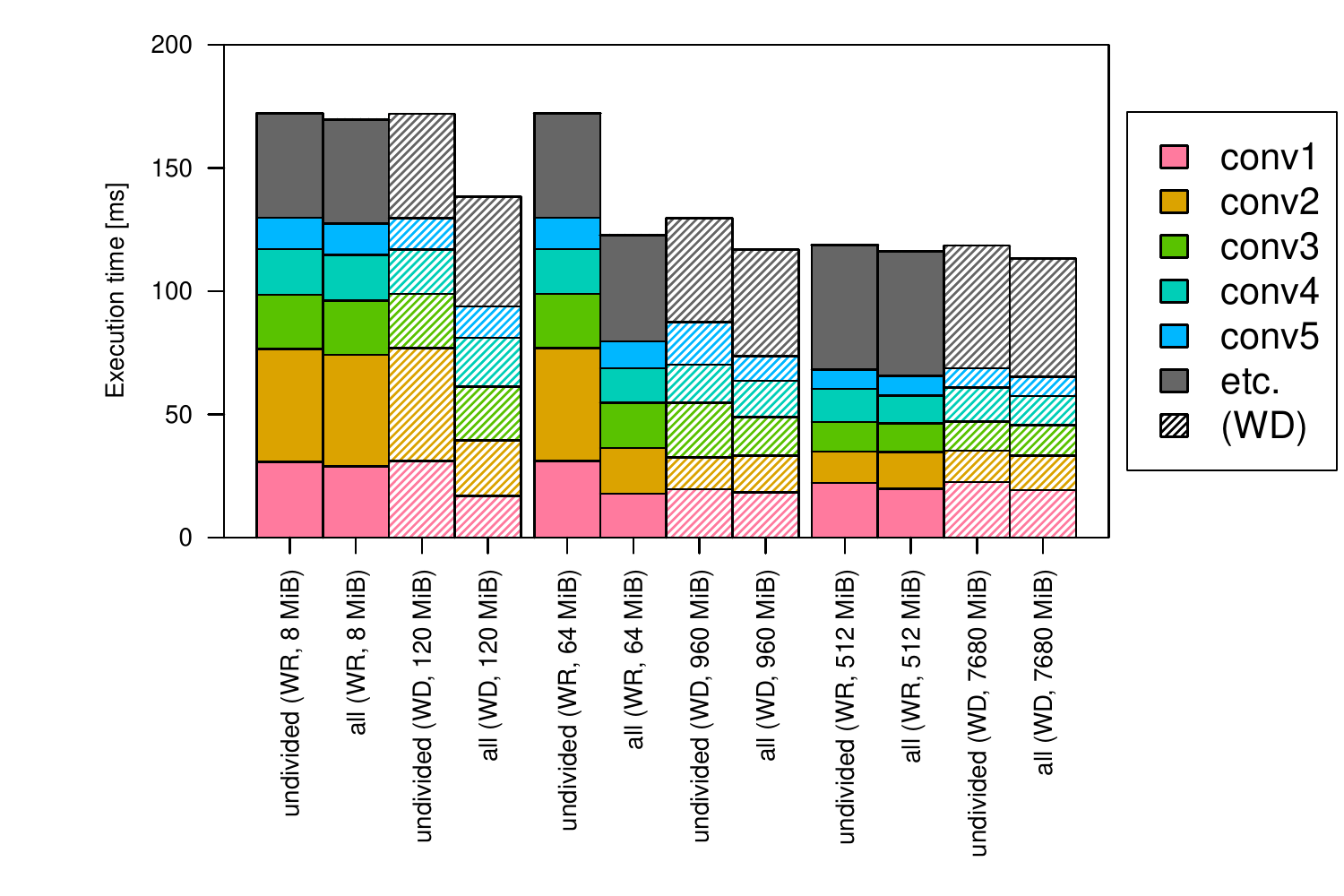}}
  \newline
  \subfloat[\resnet-50]{\includegraphics[width=\figwidth]{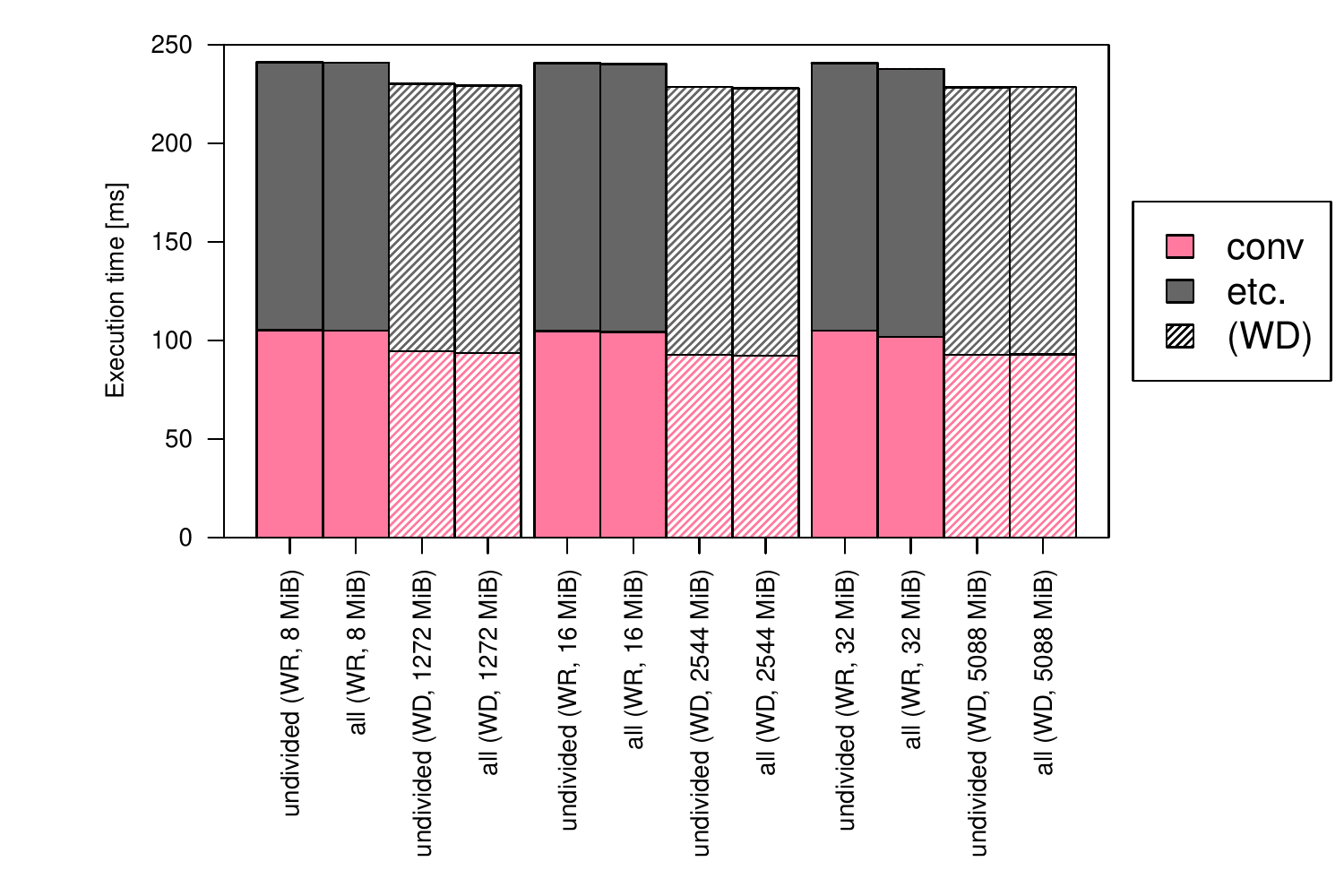}}
  \caption{Benchmark results of \alexnet \ and \resnet-50 on \phundred \ with different workspace sizes and policies (\wreuse \ and \wdiv).
    We use a mini-batch size of 256 for \alexnet \ and 32 for \resnet-50.
    Note that the adjoined bars have the same workspace limit in total.}
  \label{figure:caffe_time_wdiv}
\end{figure}

\figref{figure:caffe_time_wdiv} shows the benchmark results of using the \wdiv \ algorithm.
The adjoined bars have the same workspace limit in total: For example, since \alexnet \ has five convolutional layers and each layer has three kernels (Forward, BackwardData, BackwardFilter), we place the result with 120 MiB \wdiv \ workspace next to that of 8 MiB \wreuse \ workspaces.

\begin{figure}[t]
  \centering
  \includegraphics[width=\figwidth]{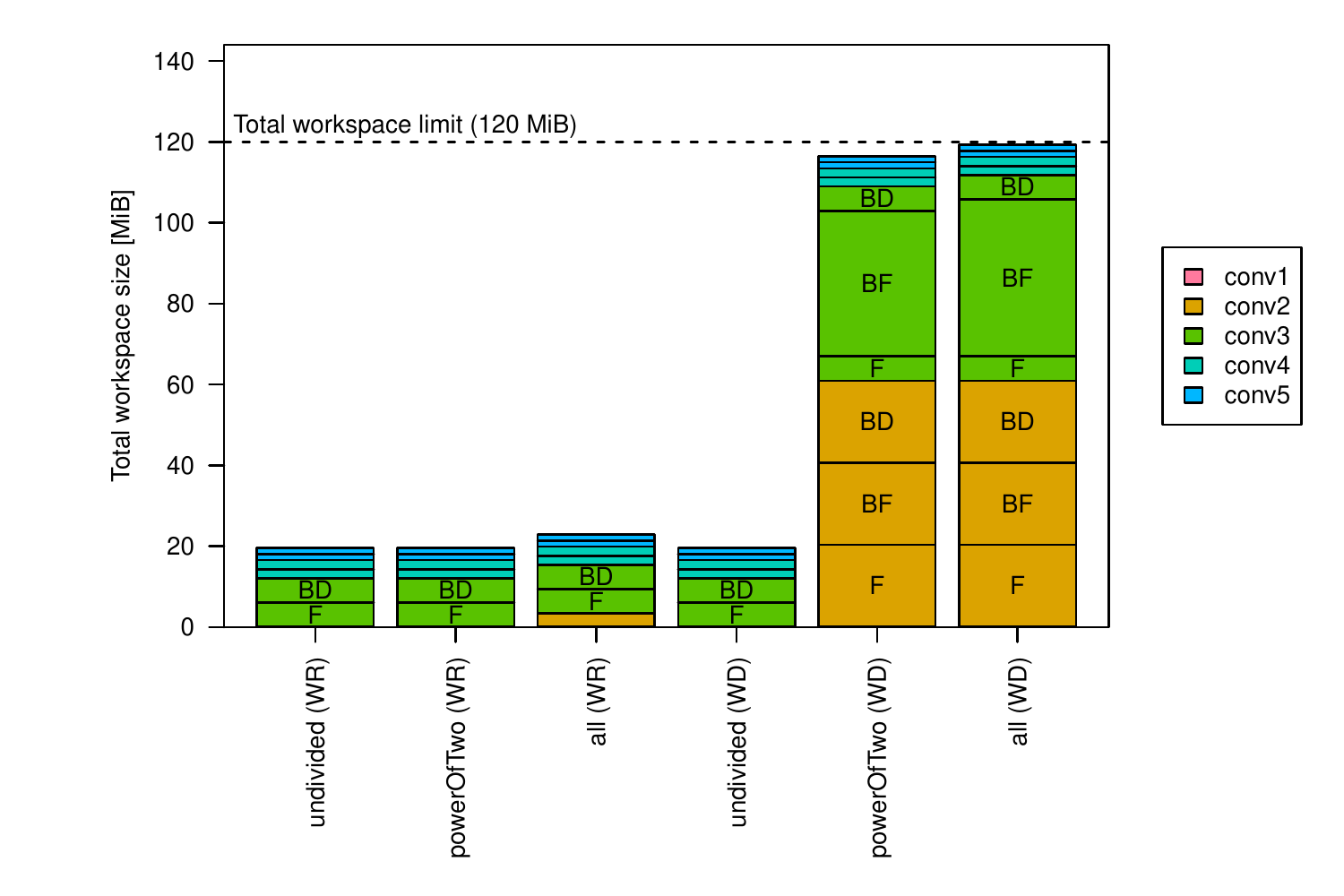}
  \caption{Assigned workspace division of \alexnet \ on \phundred.
    ``F'', ``BF'', ``BD'' represent kernel types (Forward, BackwardFilter, BackwardData respectively).
    We use a mini-batch size of 256 for \alexnet.
    We set a workspace limit of 8 MiB for \wreuse, and a total workspace limit of 120 MiB for \wdiv.}
  \label{figure:caffe_time_alexnet_ws}
\end{figure}

In \figref{figure:caffe_time_wdiv}, we can see that the training time decreases as the workspace constraints increase in both \wreuse \ and \wdiv.
At the same time, \wdiv \ successfully manages the global memory requirements better, attaining higher performance with the same overall memory footprint (see \figref{figure:caffe_time_alexnet_ws} for breakdown).
Specifically, when 120 MiB workspace in total is provided for \alexnet, the entire execution time with \wdiv \ optimization and \ttall \ option is \perf{caffe-time-wdiv-alexnet-p100-120mb-speedup}x faster than the \wreuse \ with \ttundivided \ option for the entire iteration (or \perf{caffe-time-wdiv-alexnet-p100-120mb-speedup-conv}x for convolution).
\wdiv \ also outperforms the baseline with 960 MiB workspace in total, which can use 8 times more memory for workspace, by \perf{caffe-time-wdiv-alexnet-p100-120mb-speedup-960mb-baseline}x in total execution time.

Furthermore, even for \resnet-50, which has 10 times more convolutional layers than \alexnet, \wdiv \ achieves \perf{caffe-time-wdiv-resnet50-p100-2544mb-speedup}x speedup for the entire iteration (or \perf{caffe-time-wdiv-resnet50-p100-2544mb-speedup-conv}x for convolution) with 2,544 MiB of total workspace, outperforming the original version (which consumes 5,088 MiB) in terms of memory footprint as well.
In addition, the ILP for \resnet-50 is still small enough to solve in practical time.
When the workspace limit is set to 5,088 MiB, the number of 0-1 variables is 562, and the GLPK solver takes 5.46 ms to solve it.

The main reason that \wdiv \ outperforms \wreuse \ is that in \wreuse, if \ucudnn \ fails to find better algorithms and micro-batch sizes to fully utilize the assigned workspace, \ucudnn \ must abandon that workspace slot and cannot allocate it to other kernels.
On the other hand, in \wdiv, characteristics of different desirable workspace sizes of different kernels (\figref{figure:caffe_time_alexnet_p100_smx2_tsubame3_wdiv_desirable_set_4}) are implicitly considered in the ILP-based optimization framework.
Therefore, \ucudnn \ can assign larger proportional workspaces to time-consuming layers, if it is expected that the kernels will be considerably faster with a larger workspace.

In \figref{figure:caffe_time_alexnet_ws}, \ucudnn \ with the \wdiv \ policy spares most of the workspace for ``conv2'' and ``conv3'' (\perf{caffe-time-alexnet-ws-120mb-prop-conv2-conv3}), which are the most time-consuming layers in the baseline (\wreuse, \ttundivided).
In contrast, \ucudnn \ doesn't allocate workspace of over \perf{caffe-time-alexnet-ws-120mb-ws-max-conv4-conv5} for ``conv4'' and ``conv5'', although \ucudnn \ lists some faster and desirable configurations than the baseline.
For instance, the fastest configuration of conv5 (forward), which uses FFT-based convolution with two micro-batches, is \perf{caffe-time-alexnet-ws-120mb-conv5-forward-best-speedup}x faster than baseline, although this configuration uses \perf{caffe-time-alexnet-ws-120mb-conv5-forward-best-ws} of workspace.
This observation implies that the \wdiv \ does not unnecessarily allocate workspace for a specific layer but chooses the best combination, as defined by the  ILP.

\section{Related Work}
Li et. al \cite{memeff16} propose a heuristic to automatically tune each tensor memory layout to utilize either GEMM-based or FFT-based convolution efficiently.
The proposed heuristic is, however, based on the authors' performance observation using conventional convolutional layers and specific GPU architecture,
and thus there is no guarantee that the algorithm always provides the best memory alignment for any deep neural network and GPU architecture.
On the other hand, since \ucudnn \ uses the techniques of dynamic programming and integer linear programming,
it is mathematically guaranteed that \ucudnn \ provides the best performance that the library can produce, provided that each convolution is independent from the others.

Rhu et al. \cite{7783721} propose a memory management technique that offloads neuron activations, parameters, and errors from the GPU memory to the CPU memory during forward-/backward-propagation,
so that larger models can be trained with the same memory constraint.
However, as \figref{figure:forward_memory} shows, even in such memory-efficient implementation or similar memory management techniques \cite{memreduction} \ucudnn \ is expected to save the peak memory usage of each layer.

Zlateski et al. \cite{7877151} propose ZNNi, an FFT-based convolution algorithm, and mention micro-batching technique to reduce the temporal memory usage by FFT.
\ucudnn, however, generalizes the schema so that micro-batching can be applied to any convolution algorithm, obtaining the best computational performance for the given layer configurations,
as well as maintains high portability between different existing deep learning frameworks.

\section{Conclusion}
In this paper, we proposed \ucudnn, a wrapper library for \cudnn , which divides the mini-batch to utilize high-performance convolution algorithms with limited amount of memory for workspace.
We have shown that \ucudnn \ works well even with recent CNNs, which are composed of many convolutional layers, and can easily be integrated into existing deep learning frameworks.

The performance of \ucudnn \ demonstrated in our work suggests that other layer types can be optimized as well, if they can be decomposed and computed by different algorithms. This is because \ucudnn \ does not use any special properties of the convolution operator, apart from gradient accumulation.

In addition, the result of \wdiv \ optimization (\figref{figure:caffe_time_alexnet_ws}) provides us with the insight that allocating the same workspace memory for each convolutional layer is not necessarily effective, and dynamic, adaptive assignment performs better.
This observation should be beneficial for advanced deep learning frameworks that dynamically manage GPU memory to store tensors such as neuron data, weights and their gradients, for further memory optimization.



\section*{Acknowledgment}
This research was supported by the ETH Postdoctoral Fellowship (for T. B. N.), Student Summer Research Fellowship (for Y. O.), and JST CREST Grant Number JPMJCR1303, JPMJCR1687, Japan.
Part of this work is conducted as research activities of AIST - TokyoTech Real World Big-Data Computation Open Innovation Laboratory (RWBC-OIL).



\bibliographystyle{IEEEtran}
\bibliography{IEEEabrv,references}

\end{document}